\documentclass{article}

\usepackage{microtype}
\usepackage{graphicx}
\usepackage{subcaption}
\usepackage{booktabs} 

\usepackage{hyperref}

\usepackage[accepted]{icml2024}

\usepackage{amsmath}
\usepackage{amssymb}
\usepackage{mathtools}
\usepackage{amsthm}

\usepackage[capitalize,noabbrev]{cleveref}

\usepackage{dsfont}
\usepackage{amsmath}
\usepackage{amsthm}
\usepackage{units}
\usepackage{mathtools}
\usepackage{color}

\usepackage{verbatim}
\usepackage{bm}
\usepackage{enumitem}
\usepackage[utf8]{inputenc} 
\usepackage{multirow} 
\theoremstyle{plain}

\theoremstyle{definition}

\theoremstyle{remark}

\usepackage{wrapfig}
\usepackage{soul}

\usepackage[textsize=tiny]{todonotes}

\icmltitlerunning{Coarse-to-Fine Highlighting: Reducing Knowledge Hallucination in Large Language Models}

\begin{document}

\twocolumn[
\icmltitle{Coarse-to-Fine Highlighting: Reducing 
\\Knowledge Hallucination in Large Language Models}




\begin{icmlauthorlist}
\icmlauthor{Qitan Lv}{yyy}
\icmlauthor{Jie Wang}{yyy}\hspace{-1.5mm}\textsuperscript{$\ast$}
\icmlauthor{Hanzhu Chen}{yyy}
\icmlauthor{Bin Li}{yyy}
\icmlauthor{Yongdong Zhang}{yyy}
\icmlauthor{Feng Wu}{yyy}
\end{icmlauthorlist}

\icmlaffiliation{yyy}{University of Science and Technology of China}

\icmlcorrespondingauthor{Jie Wang}{jiewangx@ustc.edu.cn}

\icmlkeywords{Machine Learning, ICML}

\vskip 0.3in
]



\begin{abstract}
Generation of plausible but incorrect factual information, often termed hallucination, has attracted significant research interest. Retrieval-augmented language model (RALM)---which enhances models with up-to-date knowledge---emerges as a promising method to reduce hallucination.
However, existing RALMs may instead exacerbate hallucination when retrieving lengthy contexts.
To address this challenge, we propose COFT, a novel \textbf{CO}arse-to-\textbf{F}ine highligh\textbf{T}ing method to focus on different granularity-level key texts, thereby avoiding getting lost in lengthy contexts. Specifically, COFT consists of three components: \textit{recaller}, 
\textit{scorer}, and \textit{selector}. First, \textit{recaller} applies a knowledge graph to extract potential key entities in a given context. Second, \textit{scorer} measures the importance of each entity by calculating its contextual weight. Finally, \textit{selector} selects high contextual weight entities with a dynamic threshold algorithm and highlights the corresponding paragraphs, sentences, or words in a coarse-to-fine manner.
Extensive experiments on the knowledge hallucination benchmark demonstrate the effectiveness of COFT, leading to a superior performance over $30\%$ in the F1 score metric.
Moreover, COFT also exhibits remarkable versatility across various long-form tasks, such as reading comprehension and question answering.

\end{abstract}

\begin{figure}[t]
    \centering 
    \includegraphics[width=1\columnwidth]{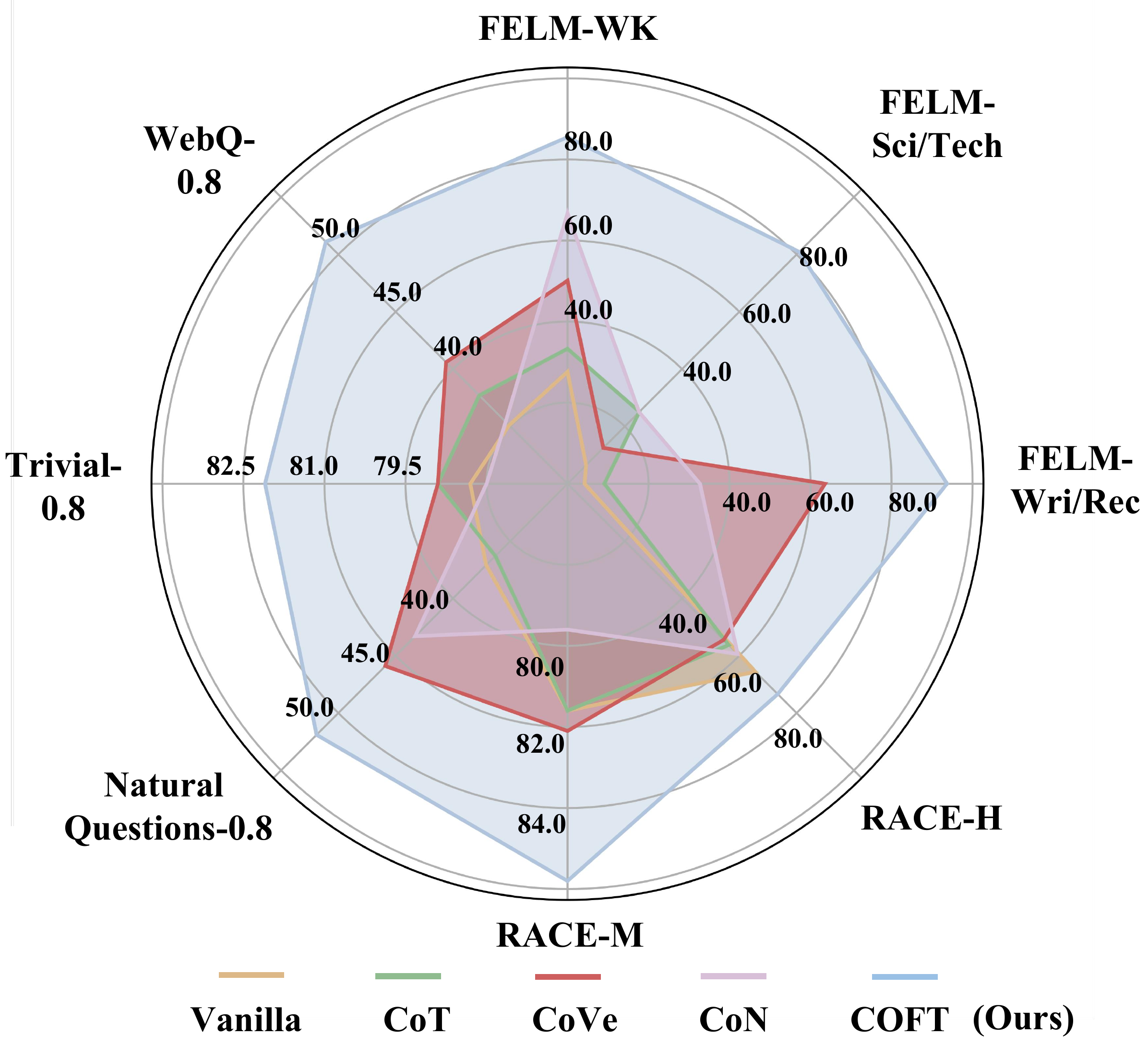}
    \caption{COFT achieves state-of-the-art performance on a broad
range of long-form tasks compared with existing methods, using ChatGPT as the backbone.}
    \label{fig:score_overview}
    \vspace{-8mm}
\end{figure}

\section{Introduction}

Large language models (LLMs) have exhibited remarkable power and impressive generalization capabilities across a wide range of domains \cite{few-shot, auto-sum}. However, even the currently most capable LLM still exhibits knowledge hallucination issues, i.e., GPT4\footnote{https://huggingface.co/spaces/lmsys/chatbot-arena-leaderboard} \citep{gpt4} may also generate plausible yet incorrect factual information \cite{hallu1}. Moreover, in long-form tasks consisting of multiple sentences or paragraphs, hallucination can be exacerbated \cite{exposurebias}. 

To address this challenge, extensive research efforts have been devoted to reducing knowledge hallucination in LLMs \cite{zeroshot, cov}. Canonical methods, such as chain-of-thought \cite{cot1}, encourage LLMs to first generate internal thoughts or reasoning steps before responding \cite{reason1, cot1}. 
 These methods enhance the logic of the reasoning process in LLMs, thereby implicitly reducing knowledge hallucination. Recently, retrieval-augmented language model (RALM) has emerged as a new trend to address hallucination, which enhances up-to-date knowledge in a plug-and-play manner \cite{search1, search2}.   Extensive works demonstrate the effectiveness of RALMs \cite{retri_survey, search2}. RALMs retrieve the most relevant contexts from external knowledge sources for LLMs to make judgments. These contexts can contain thousands of tokens, such as relevant documents from search engines or database query results \cite{lostinmid}. The potential benefit of RALM is its ability to integrate relevant external knowledge, thereby enriching the LLMs' understanding of input text and generating answers based on this information. This is particularly beneficial when LLMs lack direct knowledge of a question \cite{beni_ralms}.

 Albeit with multiple benefits of RALMs, they confront significant challenges that severely hinder their performance and deployment. 
 On the one hand, \textbf{the lack of complete contextual semantics.} When only retrieving several relevant sentences, the lack of complete contextual semantics may lead to misunderstandings. On the other hand, \textbf{the lost in the long context.} When retrieving the entire document for comprehensive information, irrelevant texts also distract their reasoning \cite{irrelevant}.
 Despite LLMs' ability to process long contexts, performances significantly decrease as the input grows longer, even for models  explicitly designed for long contexts \cite{lostinmid}.

 Therefore, in this paper, we seek to answer the question: \textit{Can we propose a novel approach that preserves complete contextual semantics and exhibits robustness to long context?} With this consideration, we delve explicitly into the two significant challenges and propose a novel approach, namely \textbf{CO}arse-to-\textbf{F}ine highligh\textbf{T}ing (COFT), which preserves complete contextual semantics and avoids getting lost in long context.
The key idea of COFT is to focus on the key texts when retrieving the entire document. COFT is a novel framework and effectively addresses the challenges within canonical RALM methods. 
Specifically, COFT consists of three components:

\begin{enumerate}[label=(\roman*), itemsep=0pt,parsep=0.1pt,topsep=0pt,partopsep=0pt]

\item \textit{Recaller} integrates an external open-source knowledge graph (KG), wikidata, to extract potential key entities as candidates within the query and reference context. To enrich the candidates, \textit{recaller} also retrieves their one-hop neighbors from the KG. 
The objective of \textit{recaller} is to identify potential key entities.

\item \textit{Scorer} applies a small language model, Llama 7B \cite{llama}, to calculate \textit{contextual weight} of each candidate entities. Entities with higher contextual weights indicate a stronger correlation with the query, and vice versa. \textit{Scorer} assigns different weights to measure the importance of each entity.

\item \textit{Selector} proposes a dynamic threshold algorithm that considers both the length and informativeness of reference contexts to select high contextual weight entities. \textit{Selector} then highlights each context based on these entities in a coarse-to-fine manner. \textit{Selector} selects the final key entities and highlights the reference context.

\end{enumerate}

COFT is a novel framework to reduce knowledge hallucination in LLMs. As shown in Figure \ref{fig:score_overview}, experiments on the knowledge
hallucination benchmark demonstrate the effectiveness of COFT with an average improvement of
$32.1\%$  in the F1 score metric. 
COFT also serves as a plug-and-play framework for many long-form tasks, which achieves an average improvement of $4.6\%$ in the precision metric for reading comprehension and a maximum improvement of $10.5\%$ in the F1 score metric for question answering.


\section{Related Work}

\begin{figure*}[t]
    \centering 
    \includegraphics[width=2\columnwidth]{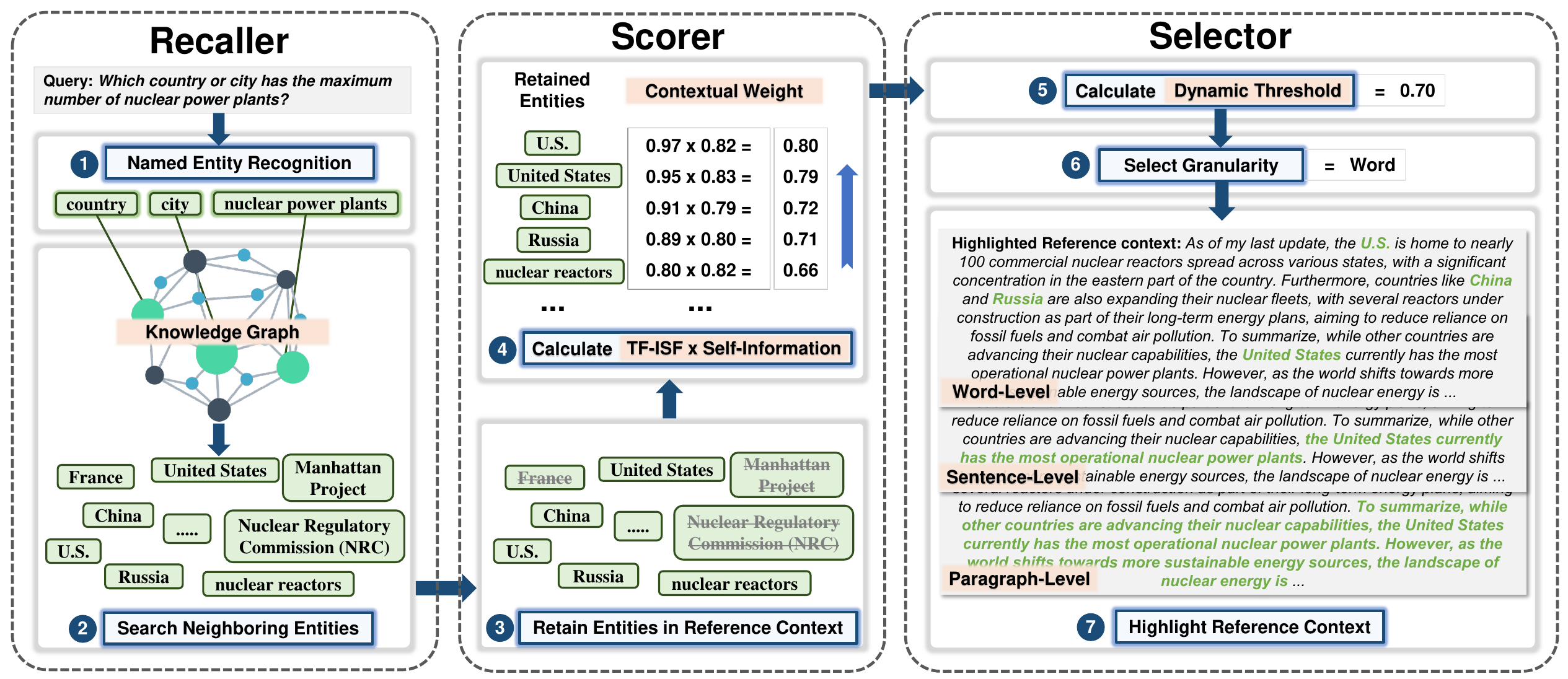}
    \caption{An overview of COFT. COFT integrates \textit{recaller}\label{model_overview}, \textit{scorer}, and \textit{selector} into a unified framework to reduce knowledge hallucination. The workflow is as follows. (1) Perform Named Entity Recognition on the query to extract potential candidate entities. (2) Search the neighboring entities for each potential entity in the knowledge graph to enrich the candidates. (3) Retain candidates that are also present in the reference context as the final key entities. (4) Calculate the contextual weight for each key entity. (5) Calculate the threshold  to filter a dynamic proportion of entities. (6) Choose the granularity for highlighting, such as word, sentence, or paragraph. (7) Highlight the reference context based on filtered entities and selected granularity.}
    \label{fig:overview}
    \vspace{-5mm}
\end{figure*}

\subsection{Retrieval-Augmented Language Models}

Retrieval-Augmented Language Models (RALMs) that enhance models with up-to-date knowledge by external knowledge sources, extend the knowledge boundaries of LLMs~\cite{1ralms,ralms_nlp,few_ralms}. These models first retrieve an external evidence corpus, such as Wikipedia, to pinpoint documents relevant to the query as reference texts~\cite{dense,que_ralm}. Then, a reader component analyzes these documents and provides a response ~\cite{lever_reader,kg_ralms}. This approach effectively retrieves reference texts related to the query, thereby enhancing the credibility of generated questions \cite{retri_survey,Longllmlingua}.
The evolution also leads to the emergence and popularity of retrieval-augmented products, such as ChatGPT plugins, Langchain, and New Bing. 

\subsection{Chain-of-\textbf{X} Approaches in LLMs}

LLMs are capable of decomposing complex problems into a series of intermediate steps and generate internal
thoughts or reasoning steps before responding, known as Chain-of-Thought (CoT) prompting~\cite{cot1}. 
The CoT approach mirrors human problem-solving by breaking complex issues into smaller components, helping LLMs focus on each segment, reducing errors, and enhancing logic in reasoning \cite{selfconsist}.
Following-up works effectively extend CoT to other chain-of-X methods, including chain-of-explanation~\cite{coe} and chain-of-knowledge~\cite{cok}.
More recently, chain-of-verification~\cite{cov} prompts
LLMs to draft initial responses, plan verification questions, answer questions, and generate the final verified response, which reduces the likelihood of LLMs to misunderstand a specific concept. Chain-of-note~\cite{search2} enables LLMs to annotate retrieved documents and incorporates
them in formulating the response to enhance the robustness of LLMs.


\subsection{Knowledge Hallucination} \label{related:hallu}

Hallucination is a general problem in LLMs, affecting various natural language processing tasks, such as reading comprehension \cite{summary}, open-domain question answering \cite{opendomain}, and remains unresolved by simply enlarging training data or model size \cite{snowball}.
We mainly discuss generation-time and retrieval-augmented methods to reduce knowledge hallucination, which are most relevant to our COFT.

For generation-time correction, efforts typically improve the token generation policy to enhance the reliability of generated contents \cite{generation1, rl_llm2}. 
Some methods enable models to generate contents along with corresponding confidence scores and correct low confidence output to reduce hallucinations \cite{retri_survey}. Chain-of-\textbf{X} approaches also improve reasoning for logical tasks, which implicitly reduces hallucination. Several approaches get improved results with extended reasoning steps, such as deductive verification \cite{DeductiveVerification, checkagain} and self-verification \cite{selfcheck, truthcheck}. 

For retrieval-augmented language models (RALMs), they mitigate hallucinations by applying external retrievers to provide query-relevant references and inject up-to-date knowledge, rather than relying solely on LLMs. RALMs can decrease hallucinations by using factual documents for grounding \cite{arag,rarh}. Several methods use automatic fact-checking and regeneration \cite{checkagain} or agreement voting and attribution analysis to conduct multi-round assessments \cite{purr,rarr}. While RALMs help reduce hallucinations, they require high-quality texts. Irrelevant texts may exacerbate hallucination and performance declines as texts grow longer \cite{lostinmid,irrelevant}.

\section{Preliminaries}
\subsection{Notations}
We denote an input prompt for LLM as $\textbf{x} = (\textbf{x}^{\text{ins}}, \textbf{x}^{\text{que}}, \textbf{x}^{\text{refs}})
$, where $\textbf{x}^{\text{ins}}$ denotes the instructions for downstream tasks, $\textbf{x}^{\text{que}}$ denotes the queries, and 
 $\textbf{x}^{\text{refs}}$ denotes reference contexts. Let $\mathcal{S}=[\textbf{s}_1, \textbf{s}_2, \textbf{s}_3, \ldots]$ denote the sentence list of $\textbf{x}^{\text{refs}}$, where $\textbf{s}_i$ denote the $i$-th sentence and $\mathcal{E}=[\textbf{e}_1, \textbf{e}_2, \textbf{e}_3, \ldots]$ denote the candidate key entity list, where $\textbf{e}_k$ denote the $k$-th candidate.
 For a given entity $\textbf{e}_k$, we denote $f_{\textbf{e}_k, \textbf{s}_i}$ and $f_{\textbf{e}_k, \mathcal{S}}$ as the number of times $\textbf{e}_k$ appears in $\textbf{s}_i$ and  $\mathcal{S}$. Let $|\textbf{s}_i|$ and $|\mathcal{S}|$ denote the number of words within sentence $\textbf{s}_i$ and the reference context $\mathcal{S}$. Let $\textbf{t}_i$ denote the $i$-th token in $\textbf{x}^{\text{refs}}$, $P(\textbf{t}_{i})$ denote its output probability by a small language model $\mathcal{M}_s$, and $I(\textbf{t}_{i})$ denote the self-information of token $\textbf{t}_{i}$. Let $\oplus$ denote the concatenation of two texts.

\subsection{Self-Information}
Self-information is a fundamental concept in information theory, which quantifies the amount of information contained in a random event \cite{communication}. In natural language processing, an event can be regarded as a generation step (i.e., a token), and the distribution corresponds to its output distribution. We can obtain self-information of a token $\textbf{t}_i$ by the follow equation: 

\vspace{-5mm}
\begin{equation*}
    I(\textbf{t}_i) = -\log_2 P\left(\textbf{t}_i \mid \textbf{t}_1, \textbf{t}_2, \ldots, \textbf{t}_{i-1}\right) 
\end{equation*}
\vspace{-8mm}

where $I(\textbf{t}_i)$ denotes the self-information of token $\textbf{t}_i$ and $P(\textbf{t}_i)$ denotes its output probability. 

In information theory, self-information represents the amount of information contained in a random event. The higher the probability of a random event occurring, the lower its self-information. 
Rare events convey more information, thus having higher self-information, while common events convey less information, resulting in lower self-information. 


In natural language processing, self-information can be utilized to evaluate the informativeness in \textbf{lexical units such as words, sentences, or paragraphs.} Lexical units with higher self-information carry important information, acting as key units that determine the semantics of the context. Conversely, lexical units with lower self-information contain less information and exert a smaller impact on the semantic interpretation of the context. Some works apply self-information in creative language \cite{nlp_selfinfo} and information compression \cite{information_compressing,llmlingua}.
The self-information between two independent events exhibits an additive property as follows:

\vspace{-5.5mm}
\begin{equation} \label{eq0}
\begin{aligned}
I\left(\textbf{t}_0, \textbf{t}_1\right) & =-\log _2 P\left(\textbf{t}_0, \textbf{t}_1\right) \\
& =-\log _2 P\left(\textbf{t}_0\right) P\left(\textbf{t}_1 \mid \textbf{t}_0\right) \\
& =-\log _2 P\left(\textbf{t}_0\right)-\log _2 P\left(\textbf{t}_1 \mid \textbf{t}_0\right) \\
& =I\left(\textbf{t}_0\right) + I\left(\textbf{t}_1\right)
\end{aligned}
\end{equation}
\vspace{-5.5mm}

This means we can measure the self-information of a lexical unit by summing the self-information of its tokens.

\section{Method}

We propose a \textbf{CO}arse-to-\textbf{F}ine highligh\textbf{T}ing method (COFT) that promotes LLMs to focus on key lexical units, which preserves complete contextual semantics and avoids getting lost in long contexts. COFT can highlight different granularity-level lexical units in a coarse-to-fine manner, such as paragraphs, sentences, and words. 
COFT organically integrates three modules in a unified framework. An overview of COFT is shown in Figure \ref{fig:overview}.


\subsection{Recaller}

\textit{Recaller} first generates candidate key entities extracted from the query and then retains the candidates occurred in the reference contexts. Specifically, for a given query and reference context, the workflow of \textit{recaller} is as follows:

\begin{enumerate}[label=(\roman*), itemsep=0pt,parsep=0.1pt,topsep=0pt,partopsep=0pt]
    \item \textit{Recaller} first conducts named entity recognition on the query to extract named entities that represent keywords within the query. These entities include some specific terms and important nouns such as people, places, organizations, etc.
    
    \item After obtaining named entities, \textit{recaller} leverages them to search one-hop neighbor entities in wikidata to enrich candidate entities. The named entities and corresponding one-hop neighbors are combined to form candidate entities for the query. 

    \item \textit{Recaller} finally retains candidate entities that are also present in the reference context, forming the final candidate key entities list. 
\end{enumerate}


As shown in the left part of Figure \ref{model_overview}, given a query such as ``Which country or city has the maximum number of nuclear power plants?'', \textit{recaller} \textbf{first} performs named entity recognition to identify entities like ``country'', ``city'', and ``nuclear power plants''. \textbf{Then}, \textit{recaller} extracts one-hop neighboring entities from wikidata for each named entity, such as ``United States'' and ``France''.
\textbf{Finally}, based on these named entities and neighboring entities,  \textit{recaller} retains entities that are present in the reference context as the final candidate key entities list. For example, ``France" will not be retained because it is not in the reference context.

\subsection{Scorer}

After obtaining candidate key entities,  \textit{scorer} proceeds to assess their importance. With this desiderata, \textit{scorer} proposes an entity-level iterative algorithm based on a small language model, Llama 7B \cite{llama} to calculate the contextual weight of each entity in the context. Algorithm \ref{algorithm1} outlines the overall procedure.
\begin{algorithm}[htb]
\renewcommand{\algorithmicrequire}{\textbf{Input:}}
\renewcommand{\algorithmicensure}{\textbf{Output:}}
\caption{Pseudo code for entity-level iterative algorithm} \label{algorithm1}
\begin{algorithmic}[1]
\REQUIRE A query $\textbf{x}^{\text{que}}$, a reference context $\textbf{x}^{\text{refs}}$, a key candidate entity list $\mathcal{E}$, and a small language model $\mathcal{M}_{s}$.

\STATE Segment the reference context $\textbf{x}^{\text{refs}}$ into sentences list $\mathcal{S}=[\textbf{s}_1, \textbf{s}_2, \textbf{s}_3, \ldots]$.
\STATE Initialize the TF-ISF dictionary $\mathcal{D}_{TF\text{-}ISF}$, the self-information dictionary $\mathcal{D}_{SI}$, and the contextual weight dictionary $\mathcal{D}_{CW}$. 
\FOR{$\textbf{e}_k \textbf{ in }  \mathcal{E}$}
   \STATE Retain $\textbf{e}_k$ occurred in each reference sentence $\textbf{s}_i \in \mathcal{S}$.
   \STATE Calculate the TF-ISF score of each entity via Equation \ref{eq1} and append entities and corresponding TF-ISF scores into $\mathcal{D}_{TF\text{-}ISF}$.
   \STATE Calculate the self-information score of each entity by the language model $\mathcal{M}_s$ via Equation \ref{eq2} and append entities and self-information scores into $\mathcal{D}_{SI}$.
   \STATE Calculate the contextual weights of each entity using $\mathcal{D}_{TF\text{-}ISF}$ and $\mathcal{D}_{SI}$ via Equation \ref{eq3} and append all entities and their contextual weights into $\mathcal{D}_{CW}$.
\ENDFOR
\ENSURE Contextual weights dictionary $\mathcal{D}_{CW}$.
\end{algorithmic}
\end{algorithm}
\vspace{-5mm}

Specifically, we first segment reference contexts $\textbf{x}^{\text{refs}}$ into sentence list $\mathcal{S}=[\textbf{s}_1, \textbf{s}_2, \textbf{s}_3, \ldots]$. Drawing upon the TF-IDF (\textbf{T}erm \textbf{F}requency–\textbf{I}nverse \textbf{D}ocument \textbf{F}requency) algorithm \cite{tf-idf}, a well-suited text relevance assessment and text mining approach that enables the exclusion of the majority of common entities while preserving important entities. We introduce the TF-I\underline{S}F algorithm, which involves considering the TF-IDF algorithm at the \underline{\textbf{S}}entence level. For a given entity $\textbf{e}_k$ in sentence $\textbf{s}_i$, the corresponding TF-ISF calculation
function is as follows:

\vspace{-5mm}
\begin{equation} \label{eq1}
TF\text{-}ISF(\textbf{e}_k) = \frac{f_{\textbf{e}_k, \textbf{s}_i}}{|\textbf{s}_i|} \times \log_2 \left( \frac{|\mathcal{S}|}{f_{\textbf{e}_k, \mathcal{S}} + 1} \right)
\end{equation}

\vspace{-3mm}

where $f_{\textbf{e}_k, \textbf{s}_i}$ and $f_{\textbf{e}_k, \mathcal{S}}$ denote the number of times $\textbf{e}_k$ appears in $\textbf{s}_i$ and $\mathcal{S}$. $|\textbf{s}_i|$ and $|\mathcal{S}|$ denote the number of words within sentence $\textbf{s}_i$ and reference contexts $\mathcal{S}$.



TF-ISF evaluates the importance of entities in reference context based on word frequency and  effectively distinguishes common but unimportant entities. Higher TF-ISF suggests that the entity plays a more important role in understanding the sentence semantics, and vice versa.

We further concatenate the query and reference context to measure the importance of each token in the reference context based on self-information. Given the input $\textbf{x}^{\text{que}} \oplus \textbf{x}^{\text{refs}}$, the self-information calculation function is as follows: 

\vspace{-6mm}
\begin{equation} \label{eq2}
I(\textbf{t}_{i}) = -\log_2 P\left( \textbf{t}_{i} \mid \textbf{x}^{\text{que}}, \textbf{t}_{1}, \textbf{t}_{2}, \ldots, \textbf{t}_{i-1} \right)
\end{equation}

\vspace{-3mm}

where $\textbf{t}_i$ denotes the $i$-th token within the reference context $\textbf{x}^{\text{refs}}$, $P(\textbf{t}_{i}| \textbf{x}^{\text{que}}, \textbf{t}_{1}, \textbf{t}_{2}, \ldots, \textbf{t}_{i-1})$ denotes its output probability by the small language model $\mathcal{M}_s$, and $I(\textbf{t}_{i})$ denotes the self-information of token $\textbf{t}_{i}$. We can further leverage the additivity property of self-information in Equation \ref{eq0} to merge tokens into entity \textbf{e}, thereby obtaining the self-information of each individual key candidate entity $I(\textbf{e})$.

To comprehensively consider both the TF-ISF and self-information, we propose contextual weights to indicate the importance of each key candidate entity in the reference context. A higher contextual weight suggests greater importance of the entity to answer the query. The contextual weight calculation function is as follows: 

\vspace{-5mm}
\begin{equation} \label{eq3}
    w(\textbf{e}_{k})= TF\text{-}ISF(\textbf{e}_{k}) \times I(\textbf{e}_{k})
\end{equation}
\vspace{-7mm}

where $TF\text{-}ISF(\textbf{e}_{k})$ and $I(\textbf{e}_{k})$ denote the TF-ISF and self-information of a key candidate entity $\textbf{e}_{k}$, respectively. Other combination methods for TF-ISF and self-information scores are also feasible, and we leave it as a future work.

\begin{table*}[ht]
	\caption{Results of knowledge hallucination benchmark on WK (world knowledge), Sci/Tech (science and technology), and Wri/Rec (writing and recommendation) domains. We denote COFT at the paragraph, sentence, and word levels as COFT\(_{p}\), COFT\(_{s}\), and COFT\(_{w}\).
 The results of vanilla, CoT, and RALM methods are taken from FELM \cite{felm}. We \textbf{bold} the best results for each LLM backbone.}\label{Tab:hallucination1}
	\centering
	\resizebox{2\columnwidth}{!}{
		\begin{tabular}{l l c c c c c c c c c} 
			\toprule
			& & \multicolumn{3}{c}{\textbf{WK}} & \multicolumn{3}{c}{\textbf{Sci/Tech}} & \multicolumn{3}{c}{\textbf{Wri/Rec}} \\
			\cmidrule(lr){3-5} \cmidrule(lr){6-8} \cmidrule(lr){9-11}
			Backbone & Methods & F1 Score & Precision & Recall & F1 Score & Precision & Recall & F1 Score & Precision & Recall \\
			\midrule
			 \multirow{8}{*}{Vicuna-33B} & Vanilla & 34.5 & 27.8 & 45.5 & 25.8 & 17.4 & 50.2 & 27.1 & 16.4 &78.7 \\
			 & CoT & 32.3 & 27.4 & 39.5 & 20.4 & 12.7 & 52.9 & 26.5 & 17.0 & 60.3 \\
			 & RALM & 48.7 & 45.7 & 52.1 & 34.2 & 24.7 & 55.8 & 27.1 & 16.2 & 82.8\\
			 & CoVe & 47.3 & 47.6 & 47.1 & 47.2 & 39.8 & 58.2 & 64.0 & 66.7 & 61.5 \\
			& CoN & 55.9 & 55.7 & 56.1 & 59.3 & 58.1 & 60.6 & 62.4 & 55.3 & 71.5 \\
			\cmidrule{2-11}
   			& COFT$_{p}$ & \textbf{69.3} & \textbf{71.9}& 66.9 & 67.9 & 62.9 & 73.8 & 70.4 & 66.8 & 74.4 \\
      	  & COFT$_{s}$ & 62.0 & {63.1} & 60.9 & 68.7 & \textbf{67.1}& 70.4 & 66.2& 64.7 & 67.7 \\
         & COFT$_{w}$ & {64.4} & {61.7} & \textbf{67.4} & \textbf{70.9} & {65.7} & \textbf{77.2} & \textbf{77.3} & \textbf{67.9} & \textbf{89.8} \\
			\bottomrule
		\end{tabular}
	}
\vspace{2mm}

	\centering
	\resizebox{2\columnwidth}{!}{
		\begin{tabular}{l l c c c c c c c c c} 
			\toprule
			& & \multicolumn{3}{c}{\textbf{WK}} & \multicolumn{3}{c}{\textbf{Sci/Tech}} & \multicolumn{3}{c}{\textbf{Wri/Rec}} \\
			\cmidrule(lr){3-5} \cmidrule(lr){6-8} \cmidrule(lr){9-11}
			Backbone & Methods & F1 Score & Precision & Recall & F1 Score & Precision & Recall & F1 Score & Precision & Recall \\
			\midrule
			 \multirow{8}{*}{ChatGPT} & Vanilla & 9.1 & 27.6 & 5.4 & 4.1 & 6.5 & 2.9 & 0.7 & 4.2 &0.4 \\
			 & CoT & 2.6 & 33.3 & 1.4 & 4.2 & 25.1 & 2.3 & 2.7 & 9.1 & 1.6 \\
			 & RALM & 25.2 & 34.9 & 19.7 & 17.4 & 16.7 & 18.2 & 20.1 & 54.1 & 12.4\\
			 & CoVe & 20.0 & 50.1 & 12.5 & 18.2 & 12.5 & 33.3 & 23.1 & 63.6 & 14.1 \\
			& CoN & 18.2 & 66.7 & 10.6 & 20.0 & 25.0 & 16.7 & 31.4 & 32.7 & 30.3\\
			\cmidrule{2-11}
      			& COFT$_{p}$ & 78.6 & 83.8& 74.0 & 83.9 & \textbf{81.2} & 86.8& 77.5 & 85.9 & 70.5 \\
      	  & COFT$_{s}$ & 76.8 & 75.7 &77.9 & 74.6 & 79.1& 70.5 & 76.8 & 84.4 & 70.5 \\
			& COFT$_{w}$ & \textbf{81.6} & \textbf{85.5} & \textbf{77.9} & \textbf{84.4} & {80.9} & \textbf{88.4} & \textbf{81.1}& \textbf{93.7} & \textbf{71.5} \\ 
			\bottomrule
		\end{tabular}
	}
 
\vspace{2mm}

	\centering
	\resizebox{2\columnwidth}{!}{
		\begin{tabular}{l l c c c c c c c c c} 
			\toprule
			& & \multicolumn{3}{c}{\textbf{WK}} & \multicolumn{3}{c}{\textbf{Sci/Tech}} & \multicolumn{3}{c}{\textbf{Wri/Rec}} \\
			\cmidrule(lr){3-5} \cmidrule(lr){6-8} \cmidrule(lr){9-11}
			Backbone & Methods & F1 Score & Precision & Recall & F1 Score & Precision & Recall & F1 Score & Precision & Recall \\
			\midrule
			 \multirow{8}{*}{GPT4} & Vanilla & 40.2 & 76.9 & 27.2 & 19.7 & 60.0 & 11.8 & 22.3 & 89.5 & 12.7 \\
			 & CoT & 50.2 & 79.4 & 36.7 & 25.2 & 64.0 & 15.7 & 26.2 & 89.1 & 15.4 \\
			 & RALM & 53.6 & 80.8 & 40.1 & 34.7 & 59.5 & 24.5 & 52.2 & 63.8 & 44.2 \\
			 & CoVe & 49.7 & 55.4 & 45.1 & 66.7 & 83.3 & 55.6 & 48.2 & 56.9 & 41.8 \\
			& CoN & 52.8 & 45.2 & 63.6 & 66.7 & 75.0 & 60.0 & 68.8 & 78.6 & 61.1 \\
			\cmidrule{2-11}
         & COFT$_{p}$ & 83.1 & 79.7 & \textbf{86.8} & \textbf{89.9} & 84.4 & \textbf{96.1}& \textbf{91.8} & 85.5 & \textbf{99.1} \\
         & COFT$_{s}$ & 80.0 & 92.3 & 70.6 & 76.6 & 84.9& 69.8 & 85.5& 89.2 & 82.1 \\
			& COFT$_{w}$  & \textbf{87.3} & \textbf{94.8} &  {80.9} & {77.9} & \textbf{86.0} & {71.3} & {84.7} & \textbf{92.9} & {77.9} \\ 
			\bottomrule
		\end{tabular}
	}
 \vspace{-5mm}
\end{table*}

\subsection{Selector}
After obtaining candidate key entities and their contextual weights, \textit{selector} highlights the final lexical units for the query. Specifically, \textit{selector} first sorts entities based on contextual weights, and proposes a dynamic threshold algorithm to filter a dynamic proportion of key entities. The dynamic thresholds can be defined as $\tau = 0.5 \times ({\tau}_{len} + {\tau}_{info}) $, where ${\tau}_{len}$ and ${\tau}_{info}$ denote the min-max normalized value of the length and informativeness for each reference context.
%
$\tau$ varies with the length and informativeness of the reference context, as longer and more informative reference context requires more highlights.
Then, \textit{selector} highlights the reference context according to the granularity of selected lexical units. This highlighting process is as follows:
\begin{enumerate}[label=(\roman*), itemsep=0pt,parsep=0.1pt,topsep=0pt,partopsep=0pt]

    \item Split the reference context according to the granularity of selected lexical units.
    
    \item Calculate the contextual weight of the split lexical units by summing the contextual weight of candidate key entities occurred in the split.
    
    
    \item Sort these lexical units in descending order by their contextual weight, and select the lexical units with contextual weights in the top $\tau \times 100\%$ for highlighting.
    
\end{enumerate}

After selecting the highlighted lexical units, \textit{selector} inserts special symbols around these lexical units. 
Considering the rich diversity of formatting found in publicly accessible web data, which forms a part of the pre-training corpus for LLMs, we adopt markdown syntax, particularly the \textbf{bold syntax} \textbf{(**)} as an example, to highlight important lexical units. This approach aligns with the natural occurrence of formatted text in online sources, thereby enabling the LLMs to more accurately interpret and process textual emphasis as it appears in real-world scenarios. 
Take word-level granularity highlighting as an example. If the selected highlighted entities are ``nuclear power plants'' and ``United States'', then the sentence ``The nuclear power plants in the United States play a crucial role in providing $\ldots$'' will be highlighted as ``The \textbf{**nuclear power plants**} in the \textbf{**United States**} play a crucial role in providing $\ldots$'' as input for LLM inference. Other highlighting methods, such as   HTML bold symbols or different markdown syntax are also viable options and we leave the exploration as a future work.



\begin{figure*}[t]
    \centering
    \begin{minipage}{1\textwidth}
        \includegraphics[width=0.33\textwidth, height=0.255\textwidth]{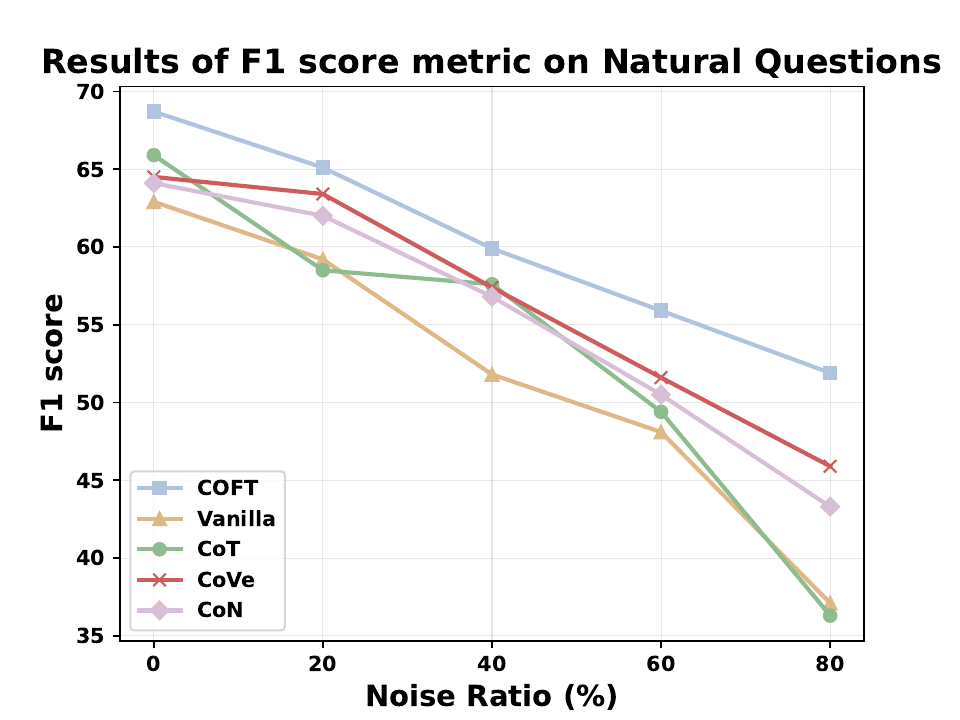}
        \includegraphics[width=0.33\textwidth, height=0.255\textwidth]{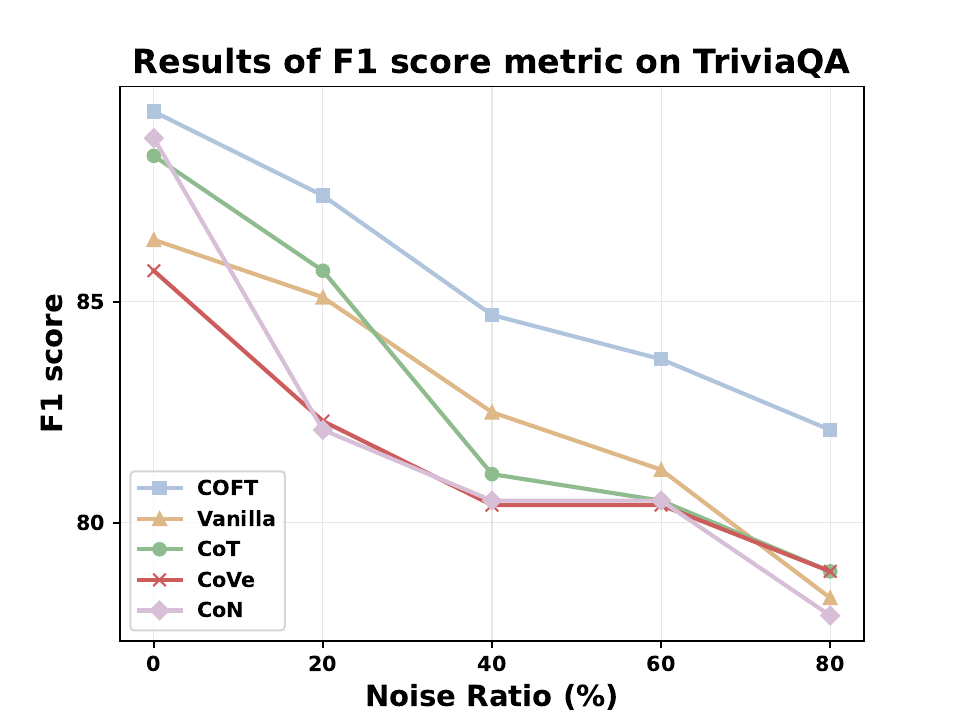}
        \includegraphics[width=0.33\textwidth, height=0.255\textwidth]{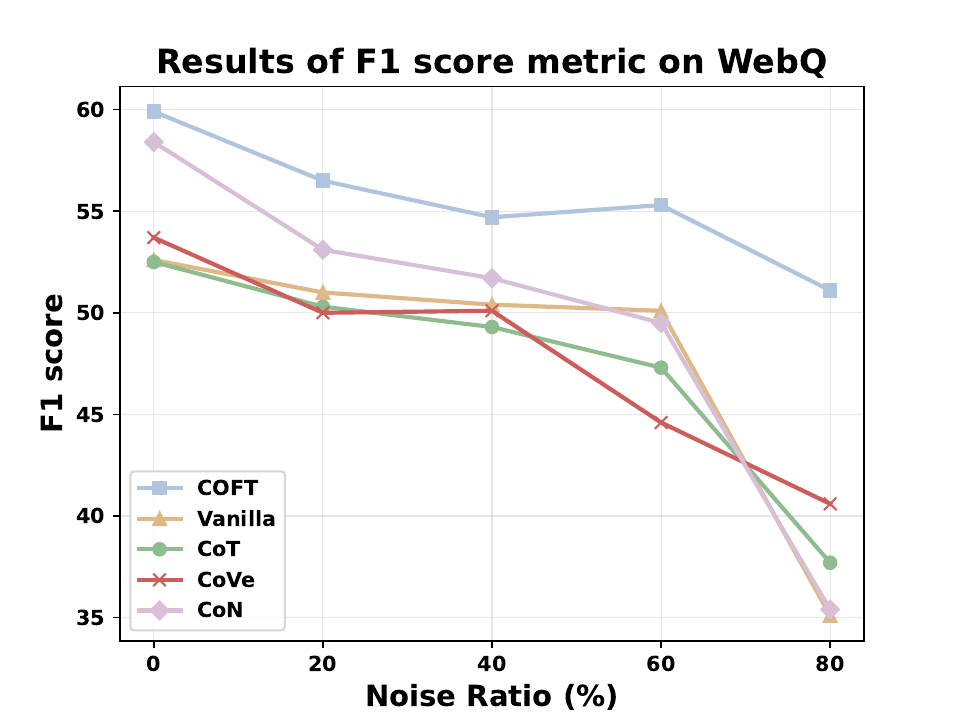}
        \caption{Evaluation on F1 score metric of noise robustness in question answering task, utilizing ChatGPT as the backbone model. COFT demonstrates superior performance on all three open-domain QA benchmarks, especially at higher noise ratios.}
        \label{fig:chatgpt_noise_roubtness_f1}
    \end{minipage}
    \vspace{-6mm}
\end{figure*}

\section{Experiments}
We design experiments to evaluate the effectiveness of COFT for reducing knowledge hallucination and demonstrate the versatility of COFT on a variety of tasks. 
With this desiderata, we divide the experiments into four parts:

\begin{enumerate}[label=(\roman*), itemsep=0pt,parsep=0.1pt,topsep=0pt,partopsep=0pt]
\item To evaluate the {effectiveness} of COFT, we compare COFT with existing state-of-the-art methods for reducing knowledge hallucination. 

\item  To demonstrate the {versatility} of COFT, we conduct experiments on reading comprehension and question-answering benchmarks.

\item To investigate the contribution of each component within COFT, we conduct the {ablation study}.

\item To provide more insight into COFT, we conduct the visualization study.


\end{enumerate}
\subsection{Experiment Setups}
\noindent \textbf{Experiment Setups.} We apply LLMs including Vicuna\footnote{https://huggingface.co/lmsys/vicuna-33b-v1.3} (\texttt{vicuna-33B-v1.3}) \cite{vicuna}, ChatGPT\footnote{https://platform.openai.com/} (\texttt{gpt-3.5-turbo}) and GPT4 (\texttt{gpt-4}) \cite{gpt4} as backbone models. To guarantee stable and reproducible results, we utilize greedy decoding and set the temperature parameter as $0$ in all experiments. 
 For \textbf{knowledge hallucination}, we use FELM \cite{felm} as the benchmark with precision, recall, and F1 score as evaluation metrics \cite{felm}. 
 For \textbf{reading comprehension}, we use RACE-H (high school level reading comprehension) and RACE-M (middle school level reading comprehension) \cite{race} as benchmarks with precision as the metric \cite{deepseek,scaling}.
 For \textbf{question answering}, we use Natural Question \cite{nq}, 
TriviaQA \cite{triviaqa}, and WebQ \cite{webq} as benchmarks with EM and F1 score as metrics \cite{refs_rc1, refs_rc3}. Details of experiment setups and datasets are in Appendix \ref{details}. For knowledge hallucination, we use word, sentence, and paragraph granularity levels of COFT (denoted as COFT$_w$, COFT$_s$, and COFT$_p$). For reading comprehension and question answering, we focus specifically on the word-level COFT$_w$ (denoted as COFT).

\noindent \textbf{Baseline Methods.} We examine five variants for each of LLMs: \textbf{(i) vanilla}: standalone LLMs without any additional preprocessing modules or external retrievers. Vanilla LLMs represent the original capabilities of LLMs. \textbf{(ii) Chain-of-thought (CoT)} \cite{cot1}: LLMs are asked to first generate internal thoughts or reasoning steps before responding. \textbf{(iii) RALM}: following \cite{felm}, we use LLMs with BM25 algorithm \cite{bm25} to retrieve the most relevant texts as input to vanilla LLMs. \textbf{(iv) Chain-of-verification (CoVe)} \cite{cov}: CoVe prompts
LLMs to draft the initial response, plan verification questions, answer the question, and generate the final verified response. \textbf{(v) Chain-of-note (CoN)} \cite{search2}: enables LLMs to 
sequentially annotate the retrieved documents and incorporates
them to formulate the response.

\subsection{Knowledge Hallucination Results} \label{sec:hallu}

In this section, we conduct experiments on the knowledge hallucination benchmark. As shown in Table \ref{Tab:hallucination1}, we observe that COFT significantly and consistently outperforms existing methods on the hallucination benchmark. Specifically, for all three backbone models, COFT achieves average improvements of $34.5\%$; $33.1\%$; $28.7\%$ in the F1 score metric, $16.3\%$; $22.6\%$; $11.6\%$ in precision metric, and $30.9\%$; $35.9\%$; $28.7\%$ in recall metric for WK (world knowledge, a wide domain including movies, countries, places, and so on), Sci/Tech (Science and Technology spanning various academic disciplines such as physics, chemistry, and biology), and Wri/Rec (Writing and Recommendation, including details of some books and movies) domains. 

While methods such as CoT and CoN do not consistently enhance the performance of Vicuna-33B and ChatGPT across various datasets, COFT consistently demonstrates a superior performance over vanilla models. Notably, in the science and technology domain, COFT achieves a maximum performance enhancement of over 60\% in the F1 score metric, which effectively underscores the importance of capturing key information in the entire context. The universality of three backbone models also suggests that COFT possesses the potential across various LLMs.

\begin{table}[hbtp]
\vspace{-5mm}
    \caption{Results of the reading comprehension task in the precision metric, utilising ChatGPT as the backbone model.}\label{Tab:rc1}
    \resizebox{\columnwidth}{!}{
    \scriptsize
        \begin{tabular}{l l c c} 
            \toprule
           \quad Backbone  &  Methods  & \quad RACE-H \quad   & \quad RACE-M \quad \\
            \midrule
            \multirow{5}{*}{\quad ChatGPT } & Vanilla & 65.6 & 81.6 \\
            & CoT & 56.3 & 81.6 \\
            & CoVe & 54.5 & 82.1\\
            & CoN & 59.4 & 79.6\\
            \cmidrule{2-4}
            & COFT & \textbf{73.4} & \textbf{85.8}\\
            \bottomrule
        \end{tabular}
    }
    \vspace{-5mm}
\end{table}



\begin{table*}[htbp]
	\caption{The results of ablation study on the knowledge hallucination benchmark, FELM,  using ChatGPT as the backbone model.}\label{Tab:ab1}
	\centering
	\resizebox{2\columnwidth}{!}{
		\begin{tabular}{l l c c c c c c c c c} 
			\toprule
			& & \multicolumn{3}{c}{\textbf
   {WK}} & \multicolumn{3}{c}{\textbf{Sci/Tech}} & \multicolumn{3}{c}{\textbf{Wri/Rec}} \\
			\cmidrule(lr){3-5} \cmidrule(lr){6-8} \cmidrule(lr){9-11}
			Backbone & Methods & F1 Score & Precision & Recall & F1 Score & Precision & Recall & F1 Score & Precision & Recall \\
			\midrule
			 \multirow{4}{*}{ChatGPT} & COFT$_{w/o \hspace{1mm} recaller}$ & 74.6 & 81.0 & 69.1 & 73.9 & 80.7 & 68.2 & 63.6 & 86.1 &60.1 \\
    			 & COFT$_{w/o \hspace{1mm} TF\text{-}ISF}$ & 78.3 & 82.4& 74.7 & 78.5 & 81.8 & 75.5 & 67.1 & 84.5 & 55.7 \\
        			 & COFT$_{w/o \hspace{1mm} SI}$ & 76.9 & 80.9& 73.3 & 76.1 & 80.5 & 72.1 & 64.5 & 85.8 & 51.7 \\
			 & COFT$_{w/o \hspace{1mm} scorer}$ & 76.2 & 80.3& 72.6 & 74.6 & 85.9 & 65.8 & 60.1 & 87.2 & 45.8 \\
			 & COFT$_{w/o \hspace{1mm} selector}$ & 77.3 & 79.7 & 75.1 & 75.7 & 82.4 & 70.1 & 70.7 & 86.1 & 60.1\\
			\cmidrule{2-11}
			& COFT & \textbf{81.6} & \textbf{85.5} & \textbf{77.9} & \textbf{85.5} & \textbf{86.5} & \textbf{84.5} & \textbf{75.2 }& \textbf{88.3} & \textbf{65.4} \\ 
			\bottomrule
   
		\end{tabular}
	}
 \vspace{-5mm}
 \end{table*}

\subsection{Reading Comprehension Results} \label{sec:rc}
Reading comprehension task necessitates that LLMs answer certain questions based on the entire content, requiring the model to retain a comprehensive understanding of the complete contextual semantics. Through the reading comprehension task, we investigate COFT's ability for full-context awareness in long contexts. We conduct experiments on RACE-H and RACE-M \cite{race} and only use the provided reading passages and do not use other information from retrieval systems. Consequently, we do not include RALM as a baseline. We present the results of COFT using ChatGPT as the backbone in Table \ref{Tab:rc1}. More Results using Vicuna-33B and GPT4 as backbones are in Appendix \ref{ap:rc}. 

As shown in Table \ref{Tab:rc1}, COFT exhibits great performance on both the RACE-H and RACE-M, which outperforms the suboptimal results by $7.8\%$ and $3.7\%$ in the precision metric. We observe that COFT achieves more performance enhancement on the more
challenging and complex dataset, RACE-H. This suggests that COFT possesses potential for application in more
complex real-world scenarios. Moreover, COFT consistently yields improved results over vanilla models, which demonstrates the effectiveness of focusing on key lexical units and maintaining full context semantics. 


\subsection{Question Answering Results} \label{exp:qa}
 Question answering task requires the LLM to effectively focus on keywords and phrases within a question. Following CoN \cite{search2}, we conduct experiments on question-answering tasks to evaluate the robustness of COFT under scenarios where reference texts contain both relevant and noisy documents.
These noise documents are retrieved based on their semantic similarity to the input questions, which often contain similar but misleading information. We employ the \textit{noise ratio} to represent the extent of noisy interference under varying degrees of noise. For instance, if the top-k documents are retrieved for LLMs, then $k \times r$ represents the number of noisy documents, while $k \times (1 - r)$ indicates the number of relevant documents. For example, with a $20\%$ noise ratio and a requirement for the top-$5$ documents, $4$ would be relevant documents, and $1$ would be a noisy document. We concatenate relevant and noisy documents randomly, to mitigate position bias \cite{vicuna}. This requires LLMs to identify the most relevant information under lengthy and noisy conditions.

As illustrated in Figure \ref{fig:chatgpt_noise_roubtness_f1}, we observe that compared to other methods, COFT demonstrates relative robustness to reference texts containing noisy text, maintaining focus on highlighted key text within reference contexts.  
These results demonstrate that COFT is robust against noisy texts, exhibiting a higher tolerance for noisy information, which more closely aligns with user inputs in real-world scenarios.

\subsection{Ablation Study}\label{sec:ab}
To further investigate the contribution of each component within COFT, we conduct a series of ablation experiments on the entire framework. We select a word-level version, COFT$_w$ to conduct the ablation study. Other granularity versions of COFT including sentences or paragraphs follow a similar way. For simplicity, we denote COFT$_w$ as COFT in this section. Specifically, we denote COFT without \textit{recaller} extracting candidate key entities as COFT$_{w/o \hspace{1mm} recaller}$, COFT without the TF-ISF score as COFT$_{w/o \hspace{1mm} TF\text{-}ISF}$, COFT without the self-information score as COFT$_{w/o \hspace{1mm} SI}$, COFT without \textit{scorer} calculating the contextual weight as COFT$_{w/o \hspace{1mm} scorer}$, and COFT without dynamic threshold selecting key candidate entities as COFT$_{w/o \hspace{1mm} selector}$, respectively. We set the threshold $\tau$ to $0.5$ for COFT$_{w/o \hspace{1mm} selector}$ as an example. More detailed results are in Appendix \ref{app:more_ab1}.

We present ablation results of COFT using ChatGPT as the backbone model in Table \ref{Tab:ab1}. More Results using backbone models including Vicuna-33B and GPT4 are in Appendix \ref{app:more_ab2}.
As shown in Table \ref{Tab:ab1}, the absence of any component within COFT results in a performance
degradation of the entire framework. Notably, \textit{recaller} and \textit{scorer} have more significant impacts on the performance of COFT,  which demonstrates the importance of extracting candidate key lexical units from the reference text and ranking them based on contextual weight to reduce knowledge hallucination.

\begin{figure}[t]
    \centering 
    \includegraphics[width=\columnwidth]{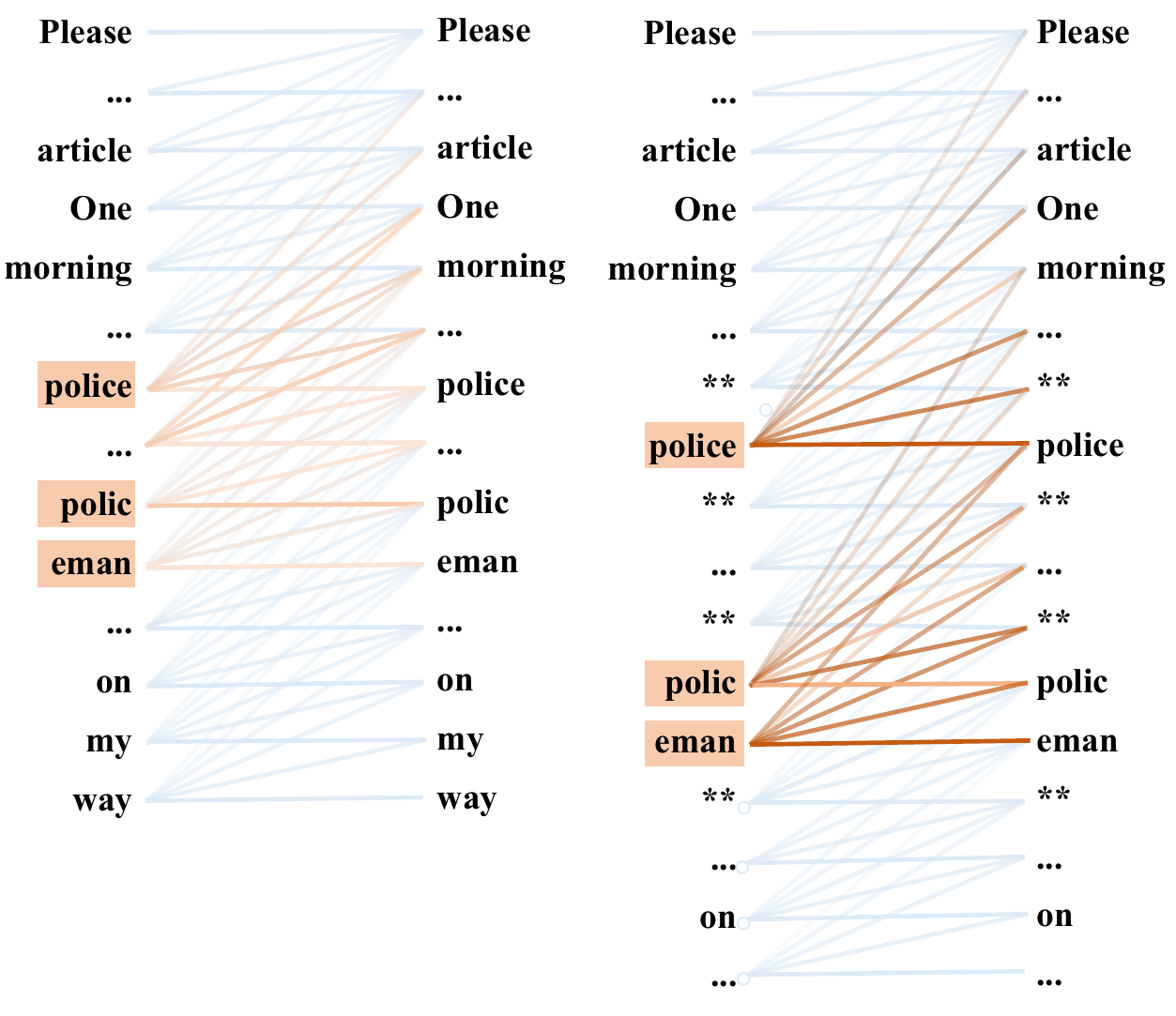}
    \caption{Visualization of the information flow in Vicuna-33B before (left) and after (right) highlighting key lexical units (between two \textbf{**} symbols). The line color depth reflects the significance of the information flow from the right word to the left.}
    \label{fig:vis_map}
    \vspace{-7mm}
\end{figure}

\subsection{Visualization Study} \label{exp:vis}
To provide more insight into COFT, we conduct a visualization study.
As mentioned above, COFT promotes LLMs to focus on key texts in the entire context. 
 We employ attention scores to trace the information flow in the reference context based on Vicuna-33B, both before and after highlighting \cite{wang2023label}. 
As shown in Figure \ref{fig:vis_map}, the highlighted key lexical units possess higher attention scores and exhibit stronger interactions with other words. This suggests that LLMs better focus on these highlighted key lexical units during inference.

\section{Conclusions}
In this paper, we propose a novel \textbf{CO}arse-to-\textbf{F}ine highligh\textbf{T}ing method to effectively reduce knowledge hallucination.
Specifically, we propose \emph{recaller}, \emph{scorer}, and \emph{selector} to form a general framework for LLMs to focus on key texts and avoid getting lost in long contexts. Extensive experiments on the knowledge
hallucination task demonstrate the effectiveness of COFT with an average improvement of
$32.1\%$ in the F1 score metric. This superior performance over existing state-of-the-art methods demonstrates the effectiveness of COFT in reducing knowledge hallucination in LLMs. COFT also serves as a plug-and-play framework for many long-form tasks that achieves an average improvement of $4.6\%$ in the precision metric for reading comprehension tasks and a maximum improvement of $10.5\%$ in the F1 score metric for question-answering tasks.


\section*{Acknowledgements}
    This work was supported in part by National Key R\&D Program of China under contract 2022ZD0119801,
    National Nature Science Foundations of China grants U23A20388, 62021001, U19B2026, and U19B2044.    
    We would like to thank all the anonymous reviewers for their insightful comments.

\section*{Impact Statements}
This paper follows existing research in the field of hallucination based on existing LLMs and open-source datasets. Moreover, our work does not involve human or animal experiments and will not provide new LLMs or datasets, so the ethical impacts and expected societal implications are those that are well established when advancing the field of hallucination.

This paper presents work whose goal is to advance the field of hallucination. There are many potential societal consequences of our work, none of which we feel must be specifically highlighted here.

As for the limitations of COFT, while COFT can significantly reduce knowledge hallucination in LLMs by focusing on key texts, it cannot update the knowledge within LLMs. Exploring low-cost methods to update knowledge within LLMs for to furture reduce  knowledge hallucination in LLMs will be the focus of our future work. We will also focus on employing this approach as a means to explicitly interpret the LLMs. 

\bibliographystyle{icml2024}

\newpage
\appendix
\onecolumn




\section{More Related Works} \label{app:more_rela}

\subsection{Language Models} \label{app:language model}
Language models such as GPT \cite{gpt1}, BERT \cite{bert}, RoBERTa \cite{roberta}, and  Megatron-LM \cite{megantron} have led to a learning paradigm shift in natural language processing (NLP). Models are first pre-trained on extensive volumes of unlabeled text corpora with language modeling objectives, and then fine-tuned on downstream tasks. Recently, large language models (LLMs) including ChatGPT, PaLM \cite{palm}, and Gemini \cite{gemini} have shown great performance in both few-shot and even zero-shot scenarios \cite{few-shot}. 

\subsection{Knowledge Hallucination}


Besides the methods mentioned in Section \ref{related:hallu} to address knowledge hallucinations during the generation time or through the RALM framework, these are some methods that address hallucinations during training time. These interventions during the training stage of LLMs to tackle the issue of model hallucinations are termed training-time correction.
For training-time correction, efforts are made to enhance the raw left-to-right outputs of either an encoder-decoder or a decoder-only language model. This enhancement involves training or suitably adjusting the model's weights to reduce the likelihood of hallucinated content. This includes using reinforcement learning \cite{rl_llm, rl_llm2} as well as contrastive learning methods \cite{Contrastive1, Contrastive2}. For training-time correction methods, models designed to resolve knowledge hallucinations during the training phase typically require the use of open-source LLMs and substantial computational resources. 
Our COFT effectively reduces the hallucination issue in LLMs without the finetuning process. Moreover, LLMs after the training-time generation method can also be integrated as a part of our COFT pipeline.

\subsection{Context Compression}
One significant challenge in the computation of self-attention mechanisms is the computational complexity $\mathcal{O}(\mathcal{L}^2)$, which exhibits a quadratic scaling in relation to the length of the input sequence $\mathcal{L}$. 
Numerous variations of the Transformer architecture have been introduced, aiming to modify the conventional attention mechanism into more efficient alternatives specifically designed for tasks involving very long context \cite{bigbird, rnntransformers}. Extensive endeavors also focus on context compression by compressing the context into fewer soft tokens.  This includes substitutes with summary tokens \cite{more1}, leveraging additional auto-encoder schemes \cite{more2}, and semantic compression \cite{more3}. Sparse attention \cite{more4} adopts a methodology predicated on learning to dynamically excise uninformative context tokens for each individual token. Several efforts also select contexts to compress the input prompt \cite{information_compressing, llmlingua, Longllmlingua}. However, due to the incomplete context, these methods may confront inevitable losses of information in real-world scenarios characterized by more complex distributions of attention. 


\section{More Details of Datasets and Experiment Setups} \label{details}

\begin{wraptable}{r}{0.6\columnwidth}
    \centering
        \caption{Statistics of the knowledge hallucination benchmark, FELM. \#Segments denotes the total number of segments. Segment Length and reference Length denote the average length of the segment and reference texts, respectively. Size denotes the number of samples for each domain.}\label{dataset:felm}
        \vspace{3mm}
    \resizebox{0.6\columnwidth}{!}{
    \begin{tabular}{l  c c c c}
    \toprule
    Dataset-Domain  & \#Segments & Segment Length & Reference Length  & Size\\
    \midrule
    FELM-WK   & 567  & 17.5 & 486.1 & 184\\
    FELM-Sci/Tech   & 717 & 19.2 & 193.6  & 125 \\
    FELM-Wri/Rec   & 1637 & 18.4 & 141.7  & 136\\
    \bottomrule
    \end{tabular}}

    \label{tab:wkqa}
\end{wraptable}

We present more details of datasets and experiment setups in this section. 

For \textbf{more details of experiment setup}, in this paper, we use ChatGPT and GPT4 as the representatives of the current closed-source LLMs, both of which can be get access via OpenAI\footnote{https://platform.openai.com/}. We apply Vicuna-33B \cite{vicuna} as a representative of open-source LLMs. All experiments were performed on four Nvidia A100 GPUs (80GB). We implement our approach based on PyTorch 1.13.0\footnote{https://pytorch.org/} and Huggingface's Transformers\footnote{https://github.com/huggingface/transformers}. For experiments with original prompts exceeding 4k tokens, we utilize extened length models, i.e.,  GPT-3.5-Turbo-16k and GPT-4-32k as our backbones. To guarantee stable and reproducible results, we utilize greedy decoding and set the temperature parameter as $0$ in all experiments. For the small language models used for calculating self-information, we apply LLaMA-7B\footnote{https://ai.meta.com/llama/}, and other open-source models can also be replaced based on specific requirements. More detailed configurations for the best performance of each task and 
 dataset can be seen within our code.

\begin{table}[ht]
\centering
\caption{Statistics of the reading comprehension benchmarks, RACE-H and RACE-M. The values below the Training/Valid/Testing Set are the number of passages and questions in each dataset, respectively. Passage/Question/Option Len denotes the average length of the passages, questions, and options, respectively. Vocab size denotes the number of words in the vocabulary.}
\label{dataset:race}
\resizebox{\columnwidth}{!}{%
\begin{tabular}{l c c c c c c c c}
\toprule
Dataset & Training Set & Valid Set & Testing Set & Passage Len & Question Len & Option Len & Vocab Size\\
\midrule
RACE-M & 6,409/25,421 & 368/1,436 & 362/1,436 & 231.1 & 9.0 & 3.9 & 32,811 \\
RACE-H & 18,728/62,445 & 1,021/3,451 & 1,045/3,498 & 353.1 & 10.4 & 5.8 & 125,120 \\
\bottomrule
\end{tabular}%
}
\end{table}
For \textbf{more details of datasets}, we list below all the datasets and corresponding evaluation metrics used in knowledge hallucination, reading comprehension, and question-answering tasks, respectively by COFT as follows.

 For the knowledge hallucination task, we employ FELM \cite{felm} as our benchmark. Specifically, FELM requires to conduct a factual evaluation of several segments based on reference texts. This requires LLMs to categorize each segment as either true or false according to the given reference context. We utilize WK (world knowledge), Sci/Tech (science/technology), and Wri/Rec (writing/recommendation) domains as our knowledge hallucination benchmark. These datasets are derived from instances where individuals prompt ChatGPT and annotators subsequently annotate the responses for factuality evaluations. We summarize the details of this knowledge hallucination benchmark in Table \ref{dataset:felm}. Following FELM \cite{felm}, we use precision, recall, and F1 score as our evaluation metrics.
    
 For the reading comprehension task, we employ RACE-M (middle school level reading comprehension task) and RACE-H (high school level reading comprehension task) \cite{race} as our benchmarks. RACE is collected from the English exams for middle and high school Chinese students in the age range between $12$ to $18$. RACE consists of nearly $28,000$ passages and nearly $100,000$ questions generated by human experts (English instructors), and covers a variety of topics that are carefully designed to evaluate the students’ ability to understand and reasoning.  The reasoning types of RACE include word matching, paraphrasing, single-sentence reasoning, multi-sentence reasoning, and insufficient/ambiguous. We summarize the details of this reading comprehension benchmark in Table \ref{dataset:race}. To better satisfy long-text reading comprehension tasks, we retain only those samples in RACE where the length of provided reading passages of top $70\%$. We also observe that the length of passages and the vocabulary size in RACE-H are significantly larger compared to RACE-M, indicating the greater difficulty level of high school examinations. For Vicuna-33B, we use one-shot setting. Following \cite{deepseek, scaling}, we use precision as our evaluation metric.

 For the question-answering task, we employ Natural Questions \cite{nq}, TriviaQA \cite{triviaqa}, and WebQ \cite{webq} as our benchmarks. We present details of these datasets as follows:

\begin{wraptable}{r}{0.5\columnwidth}
\vspace{-5mm}
    \centering
        \caption{Statistics of the question answering benchmarks.  Full size denotes the original size of these benchmarks. The IR recall evaluation is based on the retrieval of the full test set. The subset refers to the remaining dataset obtained after removing the instances that could not be retrieved. }\label{dataset:qa}
        \vspace{3mm}
    \resizebox{0.5\columnwidth}{!}{
    \begin{tabular}{l c c c}
    \toprule
    Dataset & Full Size & IR Recall & Subset Size\\
    \midrule
    Natural Questions & 3,610  &73.82  & 1,477 \\
    TriviaQA &  7,993  & 89.95 & 5,148  \\
    WebQ & 2,032  & 64.22 & 1,073\\
    \bottomrule
    \end{tabular}}

\end{wraptable}

 Natural Questions \cite{nq}: natural questions corpus comprises real anonymized, aggregated queries directed to the Google search engine. An annotator is provided with a question and a corresponding Wikipedia page from the top $5$ search results. They annotate a long answer (usually a paragraph) and a short answer (one or more entities) if they are found on the page, or they mark it as null if no long or short answer is identified. Natural Questions corpus offers a substantial dataset for end-to-end training in the field of question answering, facilitating research in natural language comprehension. It enables the study of human performance in annotating QA annotations for naturally generated questions, contributing to a better understanding of the challenges in this domain.

 TriviaQA \cite{triviaqa}: TriviaQA is a very challenging reading comprehension dataset that consists of more than 650K question-answer-evidence triples. It contains 95K question-answer pairs authored by trivia enthusiasts, accompanied by independently gathered evidence documents. On average, there are six evidence documents per question, which serve as high-quality supervision for answering the questions. TriviaQA possesses several notable characteristics: (1) It features relatively intricate and compositional questions. (2) There is substantial syntactic and lexical variability observed between questions and the corresponding answer-evidence sentences. (3) The dataset necessitates more extensive cross-sentence reasoning in order to locate answers.
 
 WebQ \cite{webq}: WebQ uses the Google Search API\footnote{https://developers.google.com/custom-search} to obtain questions that start with a specific word and contain precisely one entity. The Google Search API was employed to supply the edges of the graph. Specifically, WebQ queries the question by excluding the entity, the phrase before the entity, or the phrase after it. Each query generates five candidate questions, which are then added to the queue. This process continues until one million questions have been visited. Out of those, a random subset of $100,000$ questions is submitted to Amazon Mechanical Turk\footnote{https://www.mturk.com/} (AMT). Workers on AMT are tasked with answering the questions using only the Freebase\footnote{https://developers.google.com/freebase} page associated with the entity in the question. If the question is unanswerable based on Freebase, workers are instructed to mark it as such.

We follow the CoN \cite{search2} for text retrieval on these three datasets. 
During the process of listing retrieved documents, we set a rule to stop searching based on the number of relevant and irrelevant texts; we stop searching when both types reach our criteria. Situations wherein the DPR \cite{dense} fails to retrieve pertinent documents for certain queries will not be included in our robustness evaluation. Furthermore, to better simulate the scenarios involving long context reasoning and robustness against noisy text, we establish a criterion: for each question-answering pair, we only retain those pairs where the retrieved text exceeds $1500$ words in length. Pairs with retrieved text falling below this threshold are discarded. Consequently, the subset is more compact than the original full-size dataset set, as shown in Table \ref{dataset:qa}.




\section{More Results of Reading Comprehension} \label{ap:rc}

\begin{wraptable}{r}{0.5\textwidth} 
\vspace{-14mm}
    \caption{
Results of the reading comprehension task in the precision metric, utilizing Vicuna-33B and GPT4 as the backbone models. We \textbf{bold} the best results for each backbone, respectively.}\label{Tab:rc2}
\vspace{3mm}
    \resizebox{0.5\columnwidth}{!}{ 
    \scriptsize
        \begin{tabular}{c l c c} 
            \toprule
            \quad Backbone  &  Methods  & \quad RACE-H \quad   & \quad RACE-M \quad \\
            \midrule
            \multirow{5}{*}{\quad Vicuna-33B } & Vanilla & 44.8 & 74.2 \\
            & CoT & 43.7 & 76.6 \\
            & CoVe & 56.8 & 78.2\\
            & CoN & 51.7 & 75.7\\
            \cmidrule{2-4}
            & COFT & \textbf{68.4} & \textbf{81.3}\\

            \midrule
            \multirow{5}{*}{\quad GPT4 } & Vanilla & 78.9 & 88.4 \\
            & CoT & 87.9 & 88.6 \\
            & CoVe & 78.8 & 89.7\\
            & CoN & 79.5 & 86.2\\
            \cmidrule{2-4}
            & COFT & \textbf{89.1} & \textbf{89.9}\\
            \bottomrule
        \end{tabular}
    }
\end{wraptable}

As we mentioned above, COFT can be effectively implemented across various NLP tasks for LLM long-form inference.
In this section, we present more results of COFT with Vicuna-33B and GPT-4 as backbone models on the reading comprehension task
to serve as a supplement to Section \ref{sec:rc}, where ChatGPT is employed as the backbone model. We observe from Table \ref{Tab:rc2} that COFT consistently enhances performance across various LLM backbones in both RACE-H and RACE-M benchmarks. Specifically, COFT obtains superior performances of $11.6\%$ and $3.1\%$ in RACE-H and RACE-M for the Vicuna-33B model and $1.2\%$ and $1.3\%$ in RACE-H and RACE-M for GPT4 model, respectively. These results effectively demonstrate that COFT shows versatility under multiple LLMs as backbone models in the reading comprehension task, which also suggests that COFT effectively promotes LLMs to retain a comprehensive understanding of the long contextual semantics and to focus on keywords and phrases relevant
to the question. Furthermore, we observe that COFT achieves more performance enhancement on the more challenging and complex dataset, RACE-H. This also suggests that COFT possesses potential for application in more complex real-world scenarios. Notably, when utilizing Vicuna-33B as the backbone model, COFT achieves $11.6\%$ superior performance in precision metric on RACE-H over the suboptimal approaches. 
This also indicates the potential of COFT to better assist relatively ``small'' models in more effectively maintaining complete context semantics, focusing on key lexical units, and avoiding getting lost in the lengthy context. These findings also demonstrate the efficacy of COFT applicable in reading comprehension tasks, where complete contextual semantics are necessary.

\begin{table*}[htbp]
	\caption{The results of ablation study on the knowledge hallucination benchmark, FELM, using Vicuna-33B as the backbone model. We denote COFT without \textit{recaller} as COFT$_{w/o \hspace{1mm} recaller}$, COFT without TF-ISF score as COFT$_{w/o \hspace{1mm} TF\textit{-}ISF}$, COFT without self-information score as COFT$_{w/o \hspace{1mm} SI}$, COFT without \textit{scorer} as COFT$_{w/o \hspace{1mm} scorer}$, and COFT without \textit{selector} as COFT$_{w/o \hspace{1mm} selector}$, respectively.}\label{Tab:ab2}
 \vspace{2mm}
	\centering
	\resizebox{\columnwidth}{!}{
		\begin{tabular}{l l c c c c c c c c c} 
			\toprule
			& & \multicolumn{3}{c}{\textbf{WK}} & \multicolumn{3}{c}{\textbf{Sci/Tech}} & \multicolumn{3}{c}{\textbf{Wri/Rec}} \\
			\cmidrule(lr){3-5} \cmidrule(lr){6-8} \cmidrule(lr){9-11}
			Backbone & Methods & F1 Score & Precision & Recall & F1 Score & Precision & Recall & F1 Score & Precision & Recall \\
			\midrule
			 \multirow{6}{*}{Vicuna-33B} & COFT$_{w/o \hspace{1mm} recaller}$ & 57.8 & 55.7& 60.1 & 55.9 &60.4& 52.1 & 56.1& 54.8 &57.4 \\
    	& COFT$_{w/o \hspace{1mm} TF\text{-}ISF}$ & 60.7 & 56.5 & 65.5 & 57.9 & 55.4 & 60.7 & 64.6 & 63.5& 65.7\\
        	& COFT$_{w/o \hspace{1mm} SI}$ & 59.9 & 55.7 & 64.8 & 59.4 & 57.8 & 61.1 & 63.9 & 62.5& 65.3\\
			 & COFT$_{w/o \hspace{1mm} scorer}$ & 57.3 & 53.7 & 61.5 & 52.6 &50.8 & 54.5 &60.1 & 59.3&61.1\\
			 & COFT$_{w/o \hspace{1mm} selector}$  & 60.3 &57.9 & 62.8 & 64.7 &59.1 & 71.4&67.7 & 62.8&73.5\\
			\cmidrule{2-11}
         & COFT & \textbf{64.4} & \textbf{61.7} & \textbf{67.4} & \textbf{70.9} & \textbf{65.7} & \textbf{77.2} & \textbf{77.3} & \textbf{67.9} & \textbf{89.8} \\
			\bottomrule
		\end{tabular}
	}
 \vspace{-2mm}
 \end{table*}

 \begin{table*}[htbp]
	\caption{The results of ablation study on the knowledge hallucination benchmark, FELM using GPT4 as the backbone model. We denote COFT without \textit{recaller} as COFT$_{w/o \hspace{1mm} recaller}$, COFT without TF-ISF score as COFT$_{w/o \hspace{1mm} TF\textit{-}ISF}$, COFT without self-information score as COFT$_{w/o \hspace{1mm} SI}$, COFT without \textit{scorer} as COFT$_{w/o \hspace{1mm} scorer}$, and COFT without \textit{selector} as COFT$_{w/o \hspace{1mm} selector}$, respectively.}\label{Tab:ab3}
 \vspace{2mm}
	\centering
	\resizebox{\columnwidth}{!}{
		\begin{tabular}{l l c c c c c c c c c} 
			\toprule
			& & \multicolumn{3}{c}{\textbf{WK}} & \multicolumn{3}{c}{\textbf{Sci/Tech}} & \multicolumn{3}{c}{\textbf{Wri/Rec}} \\
			\cmidrule(lr){3-5} \cmidrule(lr){6-8} \cmidrule(lr){9-11}
			Backbone & Methods & F1 Score & Precision & Recall & F1 Score & Precision & Recall & F1 Score & Precision & Recall \\
			\midrule
			 \multirow{6}{*}{GPT4} & COFT$_{w/o \hspace{1mm} recaller}$  & 81.3 &85.6 & 77.5 & 71.9 &78.6 & 66.3 &77.1 &83.3&71.7 \\
        	& COFT$_{w/o \hspace{1mm} TF\text{-}ISF}$ & 82.7 & 89.5 & 76.9 & 74.5 & 79.2 & 70.4 & 81.5 & 86.5& 77.1\\
        	& COFT$_{w/o \hspace{1mm} SI}$ & 79.2 & 85.1 & 74.1 & 74.8 & 78.8 & 71.1 & 80.8 & 85.8& 76.3\\
			 & COFT$_{w/o \hspace{1mm} scorer}$  & 77.9 &83.4 & 73.1 & 73.4 &  76.5 & 70.6&79.4 &83.5 & 75.6\\
			 & COFT$_{w/o \hspace{1mm} selector}$  & 80.9 &84.2 & 77.9 & 74.0 & 81.8 & 67.5 &78.6 & 88.7&70.5\\
			\cmidrule{2-11}
			& COFT  & \textbf{87.3} & \textbf{94.8} &  \textbf{{80.9}} & \textbf{{77.9}} & \textbf{86.0} & \textbf{{71.3}} & \textbf{{84.5}} & \textbf{92.9} & \textbf{{77.9}} \\ 
			\bottomrule
		\end{tabular}
	}
 \end{table*}

 \begin{table*}[ht]
\centering
\caption{The inference time (per sample on average) on the FELM benchmark for the vanilla, CoT, RALM, CoVe, CoN, and COFT methods. We report the results using Vicuna-33B, ChatGPT, and GPT4 as backbone models, respectively. 
For Vicuna-33B, we deploy it locally and record the inference time. For ChatGPT and GPT4, we utilize the API interfaces provided by OpenAI to conduct inference and record the corresponding inference time. (Unit: seconds)}
\vspace{3mm}
\label{tab:inference_time}
\resizebox{0.9\columnwidth}{!}{ 
\begin{tabular}{l c c c c c c }
\toprule
Backbone Models & Vanilla & CoT & RALM & CoVe & CoN & COFT \\
\midrule
Vicuna-33B (Local Deployment) & 29.01  & 29.87 & 31.15 & 37.41 & 35.01 & 31.72 \\
ChatGPT (API) & 3.86 & 4.15 & 4.45 & 7.24 & 5.13 & 4.50 \\
GPT4 (API) & 5.14 & 5.33 & 5.74 & 10.22 & 6.68 & 5.81 \\
\bottomrule
\end{tabular}
}

\end{table*}

\begin{table*}[htbp]
	\caption{Results of question answering tasks in Natural Questions, TriviaQA, and WebQ benchmarks, utilizing ChatGPT as the backbone model. We evaluate the performance of each method in terms of EM and F1 score across various noise ratios \cite{search2}. We \textbf{bold} the best results for each noise ratio, respectively.}\label{Tab:qa2}
 \vspace{3mm}
	\centering
	\resizebox{\columnwidth}{!}{
		\begin{tabular}{l l c c c c c c c} 
			\toprule
			& & &  \multicolumn{2}{c}{\textbf{NQ}} & \multicolumn{2}{c}{\textbf{TriviaQA}}  &\multicolumn{2}{c}{\textbf{WebQ}} \\
			\cmidrule(lr){4-5} \cmidrule(lr){6-7} \cmidrule(lr){8-9}
			Backbone & Methods & Noise Ratio & EM &    F1 Score & EM & F1 Score & EM & F1 Score \\
			\midrule
			 \multirow{27}{*}{ChatGPT} & Vanilla &\multirow{5}{*}{80\%}  &25.9 &37.1 &67.4 &78.3 &13.9 &35.1 \\
			 & CoT & &24.5 &36.3 &69.6 &78.9 &13.8 &37.7\\
			 & CoVe &  &27.1 &45.9 &68.8 &78.9 &24.4 &40.6\\
			& CoN &  &23.9 &43.3 &68.5 &77.9 &17.2 &35.4\\
			& COFT  & &33.6 &51.9 &74.3 &82.1 &27.6 &51.1\\
			\cmidrule(lr){2-9}
   			& Vanilla &\multirow{5}{*}{60\%}  &37.7 &48.1 &70.4 &81.2 &33.7 &50.1 \\
			 & CoT & &36.1 &49.4 &72.2 &80.5 &34.5 &47.3\\
			
			 & CoVe &  &32.8 &51.6 &71.3 &80.4 &35.7 &44.6\\
			& CoN &  &37.7 &50.5 &68.7 &80.5 &34.5 &49.5\\
			& COFT  & &43.2 &55.9 &75.1 &83.7 &37.3 &55.3\\
			\cmidrule(lr){2-9}
   			 & Vanilla &\multirow{5}{*}{40\%}  &37.0 &51.8 &73.3 &82.5 &31.1 &50.4 \\
			 & CoT & &40.7 &57.6 &73.7 &81.1 &31.9 &49.3\\
			
			 & CoVe &  &39.3 &57.4 &72.5 &80.4 &34.7 &50.1\\
			& CoN &  &42.7 &56.8 &70.4 &80.5 &35.1 &51.7\\
			& COFT  & &46.4 & 59.9 &76.6 &84.7 &35.7 &54.7\\
			\cmidrule(lr){2-9}
   			 & Vanilla &\multirow{5}{*}{20\%}  &44.4 &59.2 &73.5 &85.1 &35.5 &51.0 \\
			 & CoT & &46.2 &58.5 &75.4 &85.7 &35.7 &50.3\\
			
			 & CoVe &  &42.1 &63.4 &76.7 &82.3 &34.4 &50.0\\
			& CoN &  &46.7 &62.0 &71.9 &82.1 &35.7 &53.1\\
			& COFT  & &50.2 &65.1 &79.2 &87.4 &38.7 &56.5\\
			\cmidrule(lr){2-9}
   			\cmidrule(lr){2-9}
   			 & Vanilla &\multirow{5}{*}{0\%}  &49.8 &62.9 &75.5 &86.4 &35.4 &52.6 \\
			 & CoT & &51.6 &65.9 &75.3 &88.3 &35.5 &52.5\\
			 & CoVe &  & 44.8 & 64.5 &75.4 &85.7 & 34.5 & 53.7\\
			& CoN &  &50.6 &64.1 &78.7 &88.7 &35.5 &58.4\\
			& COFT  &  & 53.9 &68.7 &79.2 &89.3 &38.7 &59.9\\
   \bottomrule
		\end{tabular}
	}
 \end{table*}

\begin{table*}[htbp]
	\caption{Results of question-answering tasks in Natural Questions, TriviaQA, and WebQ benchmarks, utilizing Vicuna-33B as the backbone model. We evaluate the performance of each method in terms of EM and F1 score across various noise ratios \cite{search2}. We \textbf{bold} the best results for each noise ratio, respectively.}\label{Tab:qa1}
 \vspace{3mm}
	\centering
	\resizebox{1.0\columnwidth}{!}{
		\begin{tabular}{l l c c c c c c c} 
			\toprule
			& & &  \multicolumn{2}{c}{\textbf{NQ}} & \multicolumn{2}{c}{\textbf{TriviaQA}}  &\multicolumn{2}{c}{\textbf{WebQ}} \\
			\cmidrule(lr){4-5} \cmidrule(lr){6-7} \cmidrule(lr){8-9}
			Backbone & Methods & Noise Ratio & EM & F1 Score & EM & F1 Score & EM & F1 Score \\
			\midrule
			 \multirow{27}{*}{Vicuna-33B} & Vanilla &\multirow{6}{*}{80\%}  &15.7 &22.1 &42.4 &50.1 &9.2 &13.8 \\
			 & CoT & &17.2 &25.8 &43.1 &51.5 &9.6 &15.5\\
			 & CoVe &  &16.4 &24.9 &45.2 &53.7 &11.7 &16.4\\
			& CoN &  &16.2 &24.3 &46.8 &54.9 &10.5 &15.5\\
			& COFT  & &19.7 &30.6 &49.2 &58.6 &13.8 &21.0\\
			\cmidrule(lr){2-9}
   			& Vanilla &\multirow{6}{*}{60\%}  &17.9 & 27.6 &46.8 &57.7 &15.7 &20.5 \\
			 & CoT & &17.8 &28.1 &48.5 &59.2 &13.1 &21.1\\
			 & CoVe &  &19.9 &27.9 &49.1 &61.4 &14.8 &24.1\\
			& CoN &  &17.3 &28.5 &49.5 &60.6 &15.7 &26.4\\
			& COFT  & &21.3 &32.8 &52.8 &63.1 &16.2 &28.2\\
			\cmidrule(lr){2-9}
   			 & Vanilla &\multirow{6}{*}{40\%}  &18.4 &28.5 &52.1 &63.0 &14.9&24.7 \\
			 & CoT & &18.9 &29.4 &53.5 &62.5 &14.4 &25.7\\
			 & CoVe &  &22.5 &32.1 &53.4 &62.8 &15.8 &28.4\\
			& CoN &  &20.1 &31.6 &52.8 &63.4 &15.5 &27.3\\
			& COFT  & &23.7 &35.0 &55.9 &65.8 &19.7 &30.4\\
			\cmidrule(lr){2-9}
   			 & Vanilla &\multirow{6}{*}{20\%}  &19.4 &32.6 &54.3 &63.3 &20.4 &29.8 \\
			 & CoT & &20.7 &32.9 &54.9 &63.9 &20.9 &30.5\\
			 & CoVe &  &23.5 &33.7 &55.3 &64.3 &22.5 &31.7\\
			& CoN &  &21.9 &34.3 &55.7 &63.4 &21.8 &30.2\\
			& COFT  & &25.5 &36.2 &57.3 &67.5 &22.4 &33.4\\
			\cmidrule(lr){2-9}
   			\cmidrule(lr){2-9}
   			 & Vanilla &\multirow{6}{*}{0\%}  &21.5 &34.9 &56.9 &66.5 &22.8 &32.7 \\
			 & CoT & &24.5 &35.5 &57.6 &64.2 &20.4 &31.5\\
			 & CoVe &  &25.7 &38.1 &59.4 &66.7 &25.8 &35.7\\
			& CoN &  &24.2 &37.6 &58.9 &65.8 &24.6 &36.9\\
			& COFT  & &26.7 &39.5 &59.8 &68.4 &26.5 &35.9\\
   \bottomrule
		\end{tabular}
	}
 \end{table*}

\begin{table*}[htbp]
	\caption{Results of question answering tasks in Natural Questions, TriviaQA, and WebQ benchmarks, utilizing GPT4 as the backbone model. We evaluate the performance of each method in terms of EM and F1 score across various noise ratios \cite{search2}. We \textbf{bold} the best results for each noise ratio, respectively.}\label{Tab:qa3}
 \vspace{3mm}
	\centering
	\resizebox{1.0\columnwidth}{!}{
		\begin{tabular}{l l c c c c c c c} 
			\toprule
			& & &  \multicolumn{2}{c}{\textbf{NQ}} & \multicolumn{2}{c}{\textbf{TriviaQA}}  &\multicolumn{2}{c}{\textbf{WebQ}} \\
			\cmidrule(lr){4-5} \cmidrule(lr){6-7} \cmidrule(lr){8-9}
			Backbone & Methods & Noise Ratio & EM &   F1 Score & EM & F1 Score & EM & F1 Score \\
			\midrule
			 \multirow{27}{*}{GPT4} & Vanilla &\multirow{5}{*}{80\%}  &53.3 &60.1 &45.3 &58.5 &19.7 &38.7 \\
			 & CoT & &54.1 &62.4 &49.6 &60.1 &20.7 &40.6\\
			
			 & CoVe &  &53.4 &60.9 &44.5 &59.1 &16.7 &39.3\\
			& CoN &  &54.3 &61.8 &50.8&60.4 &6.9 &31.9\\
			& COFT  & &57.3 &65.4 &54.4 &62.7 &24.6 &43.0\\
			\cmidrule(lr){2-9}
   			& Vanilla &\multirow{5}{*}{60\%}  &54.9 &63.2 & 49.8 &65.5 &20.1 &40.1 \\
			 & CoT & &57.3 &66.1 &55.2 &61.5 &20.7 &40.4\\
			 
			 & CoVe &  &54.0 &64.4 &45.4 &61.4 &20.3 &41.9\\
			& CoN &  &56.8 &66.1 &50.0 &66.8 &10.3 &33.3\\
			& COFT  & &59.6 &67.8 &57.8 &68.4 &24.1 &44.1\\
			\cmidrule(lr){2-9}
   			 & Vanilla &\multirow{5}{*}{40\%}  &57.1 &66.3 &54.8 &65.4 &19.4 &41.5 \\
			 & CoT & &57.7 &67.8 &56.7 &67.6 &22.6 &42.5\\
			
			 & CoVe &  &58.4 &66.1 &55.4 &64.1 &21.4 &41.1\\
			& CoN &  &58.2&67.1  &55.6 &66.1&16.4 &36.0\\
			& COFT  & &59.1 &70.3 &59.3 &72.4 &25.8 &44.8\\
			\cmidrule(lr){2-9}
   			 & Vanilla &\multirow{5}{*}{20\%}  &60.6 &68.8 &58.8 &75.3 &16.1 &39.5 \\
			 & CoT & &61.1 &69.4 &58.3 &72.5 &22.8 &41.8\\
			
			 & CoVe &  &58.8 &70.0 &55.7&70.9 &22.7 &39.3\\
			& CoN &  &61.4 &69.8 &57.8 &70.8 &16.5 &36.6\\
			& COFT  & &62.4 &72.1 &62.3 &74.1 &28.8 &44.6\\
			\cmidrule(lr){2-9}
   			 & Vanilla &\multirow{5}{*}{0\%}  &63.5 &74.8 &62.5 & 81.4&19.4 &42.8 \\
			 & CoT & &63.8 &74.4 &62.5 &82.1 &22.6 &41.7\\
		
			 & CoVe &  &59.6 &71.4 &62.2 &73.3 &23.3 &43.4\\
			& CoN &  &63.2 &74.1 &59.8 &76.7 &16.8 &36.2\\
			& COFT  & & 64.3 &75.7 &66.7 &82.3 &29.7 &46.2\\
   \bottomrule
		\end{tabular}
	}
 \end{table*}

\section{More Results of Question Answering}

\begin{figure*}[t]
    \centering
    \begin{minipage}{1\textwidth}
        \includegraphics[width=0.32\textwidth, height=0.255\textwidth]{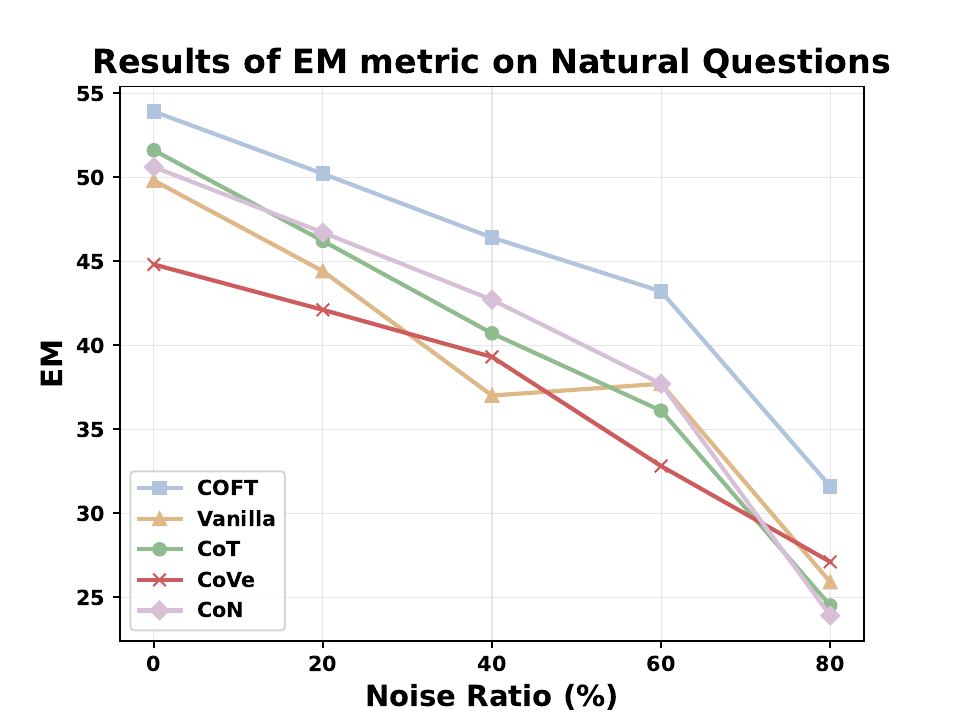}
        \includegraphics[width=0.32\textwidth, height=0.255\textwidth]{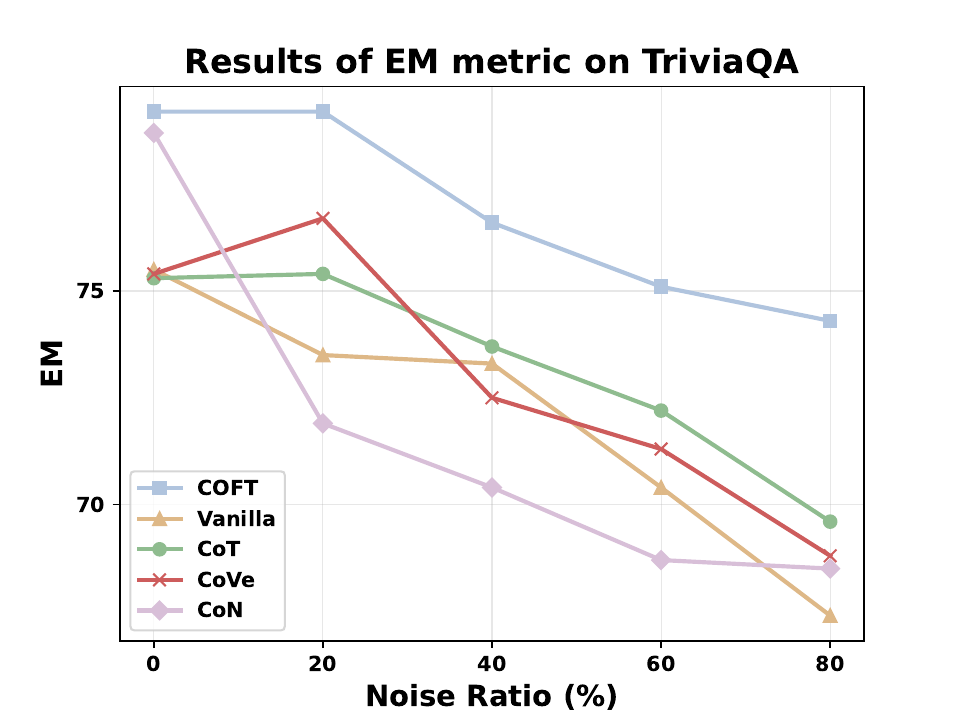}
        \includegraphics[width=0.32\textwidth, height=0.255\textwidth]{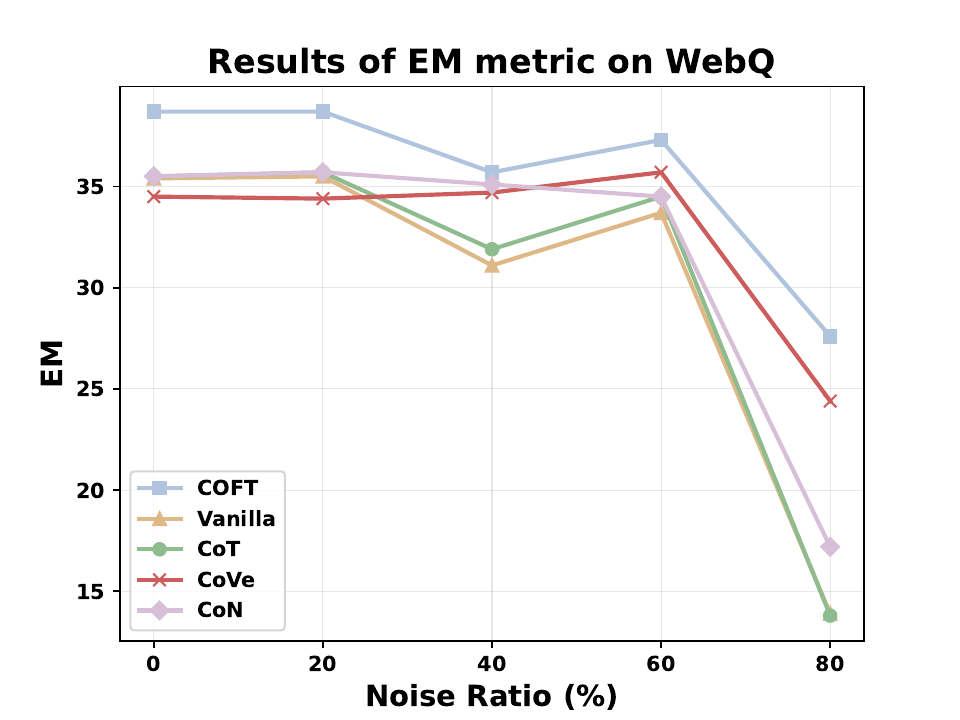}
        \caption{Evaluation on EM metric of noise robustness in question answering task, utilizing ChatGPT as the backbone model: COFT demonstrates superior performance on all three open-domain QA benchmarks, especially at higher noise ratios.}
        \label{fig:chatgpt_noise_roubtness_EM}
    \end{minipage}
\end{figure*}

\begin{figure*}[t]
    \centering
    \begin{minipage}{1\textwidth}
        \includegraphics[width=0.32\textwidth, height=0.255\textwidth]{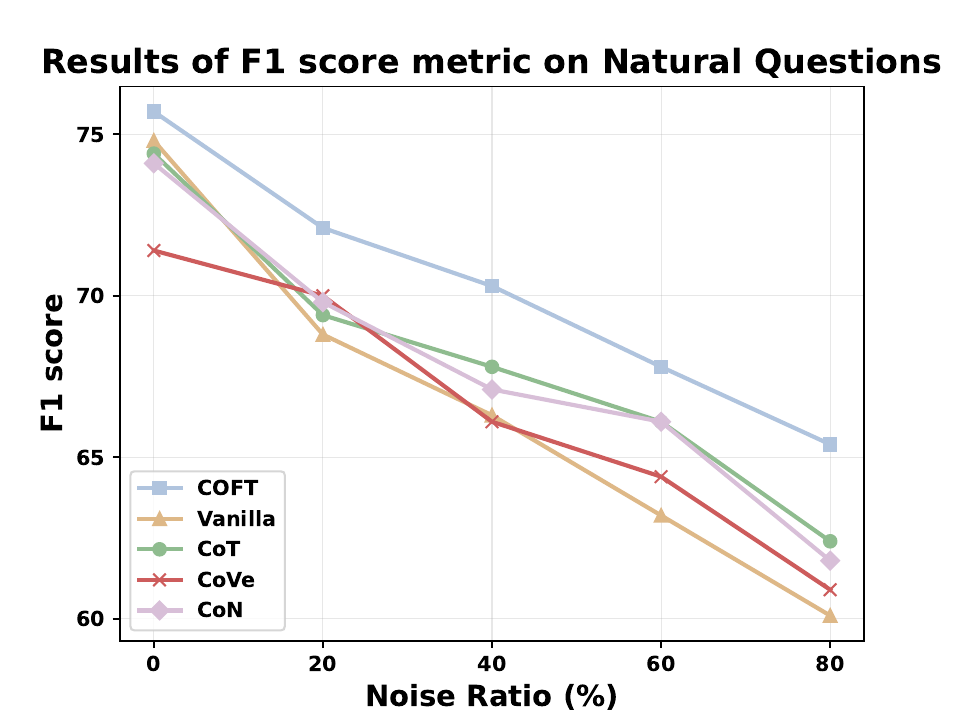}
        \includegraphics[width=0.32\textwidth, height=0.255\textwidth]{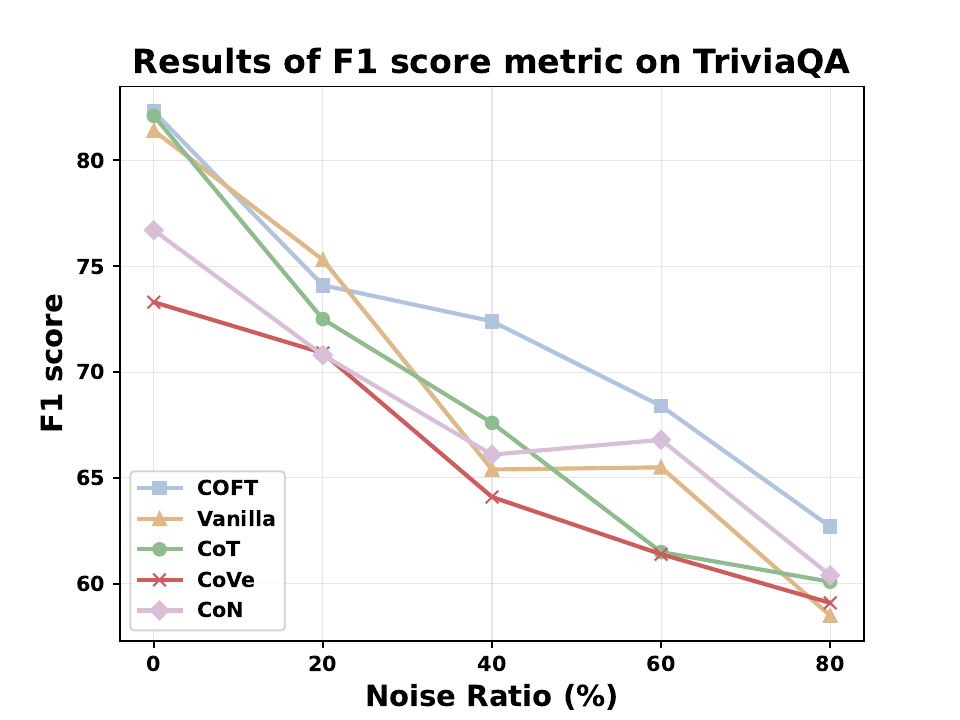}
        \includegraphics[width=0.32\textwidth, height=0.255\textwidth]{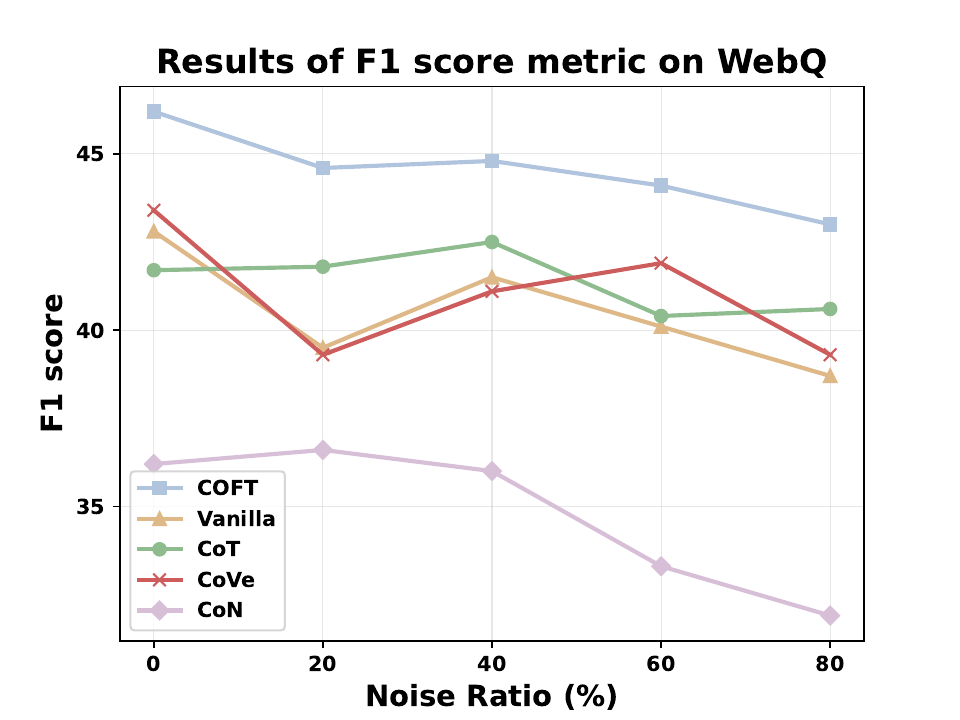}
        \caption{Evaluation on F1 score metric of noise robustness in question answering task, utilizing GPT4 as the backbone model: COFT demonstrates superior performance on all three open-domain QA benchmarks, especially at higher noise ratios.}
        \label{fig:GPT4_noise_roubtness_f1}
    \end{minipage}
\end{figure*}

\begin{figure*}[t]
    \centering
    \begin{minipage}{1\textwidth}
        \includegraphics[width=0.32\textwidth, height=0.255\textwidth]{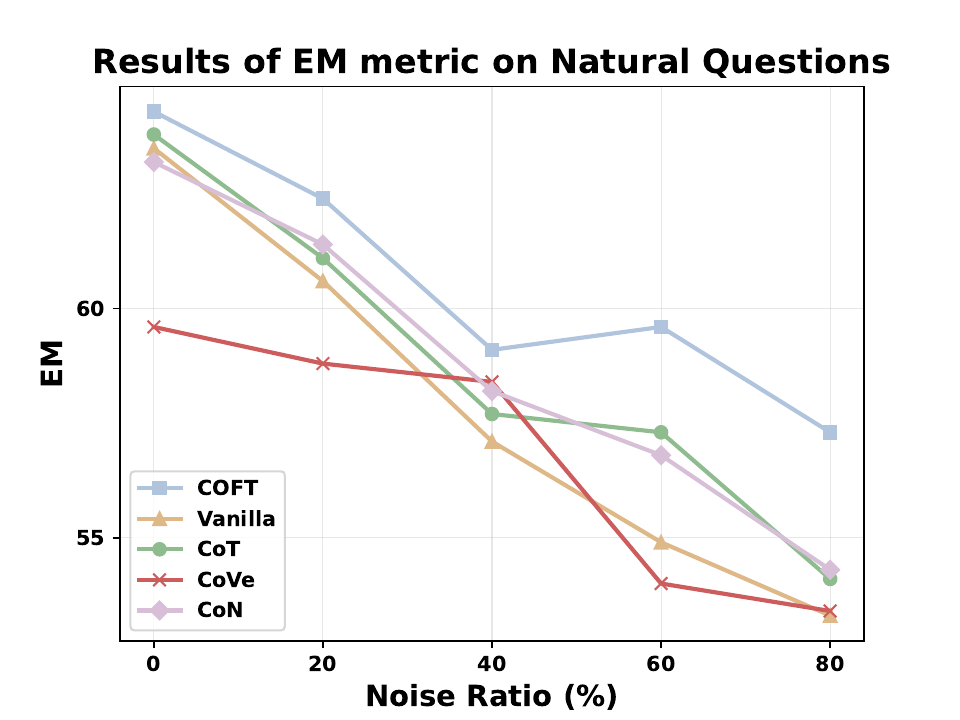}
        \includegraphics[width=0.32\textwidth, height=0.255\textwidth]{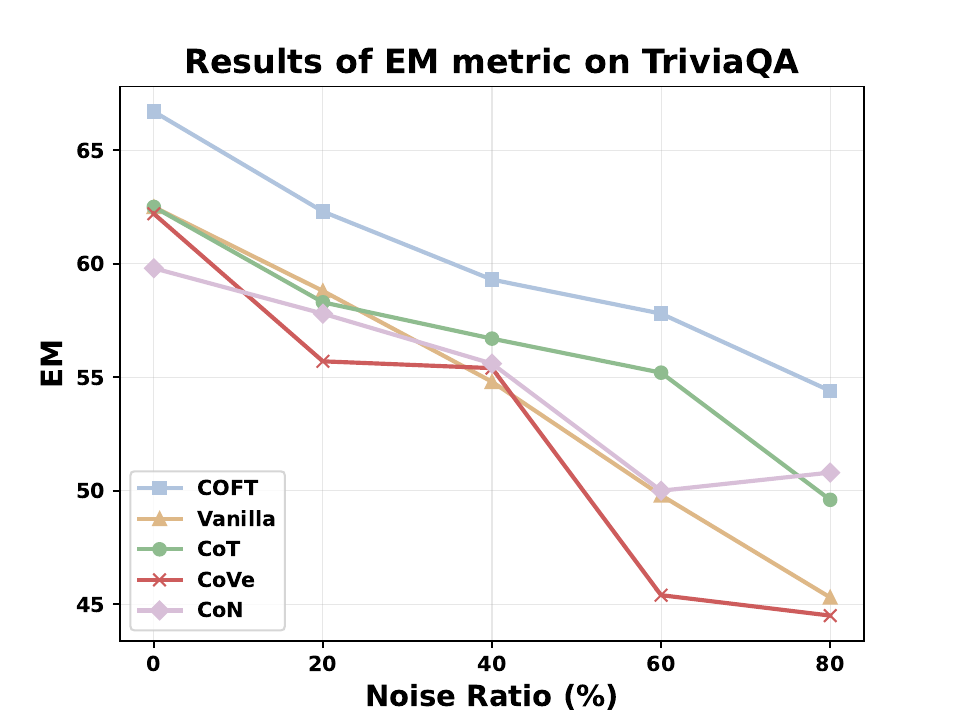}
        \includegraphics[width=0.32\textwidth, height=0.255\textwidth]{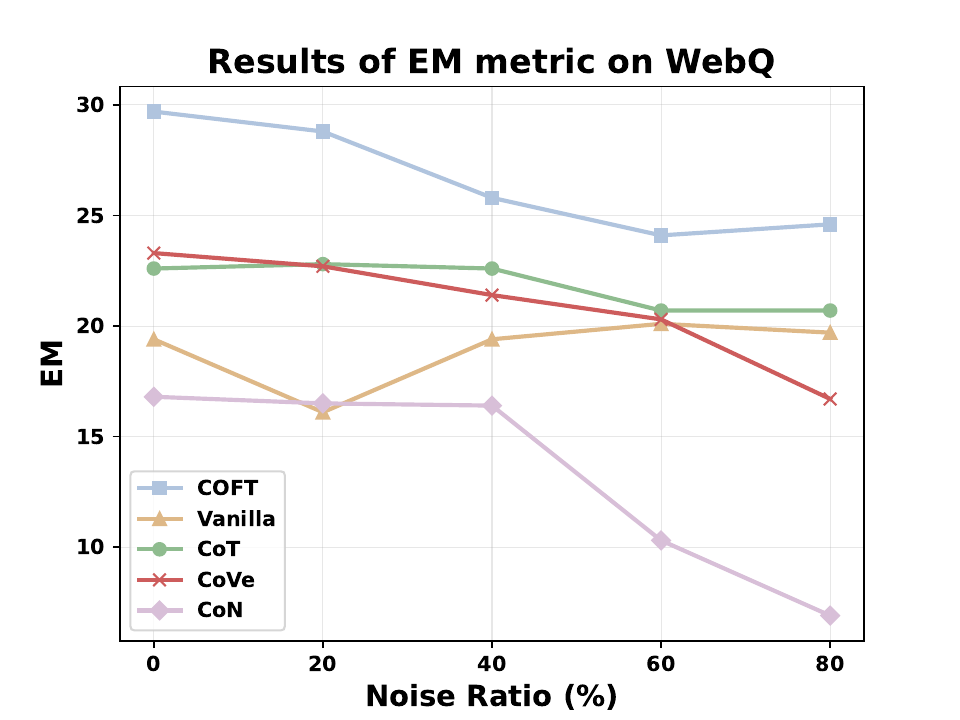}
        \caption{Evaluation on EM metric of noise robustness in question answering task, utilizing GPT4 as the backbone model: COFT demonstrates superior performance on all three open-domain QA benchmarks, especially at higher noise ratios.}
        \label{fig:GPT4_noise_roubtness_EM}
    \end{minipage}
\end{figure*}

\begin{figure*}[t]
    \centering
    \begin{minipage}{1\textwidth}
        \includegraphics[width=0.32\textwidth, height=0.255\textwidth]{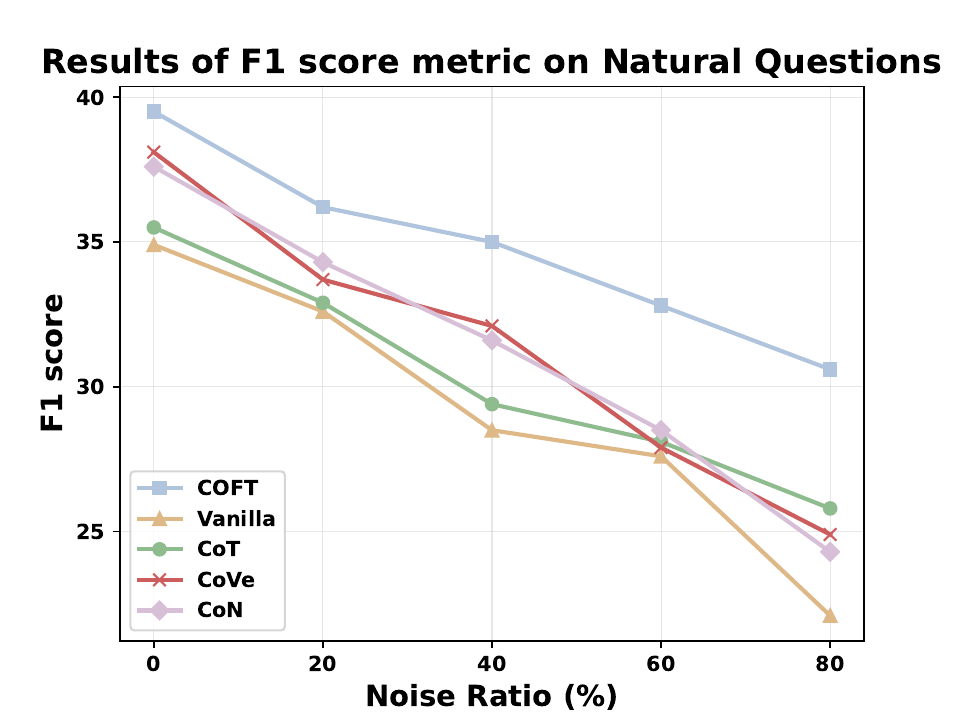}
        \includegraphics[width=0.32\textwidth, height=0.255\textwidth]{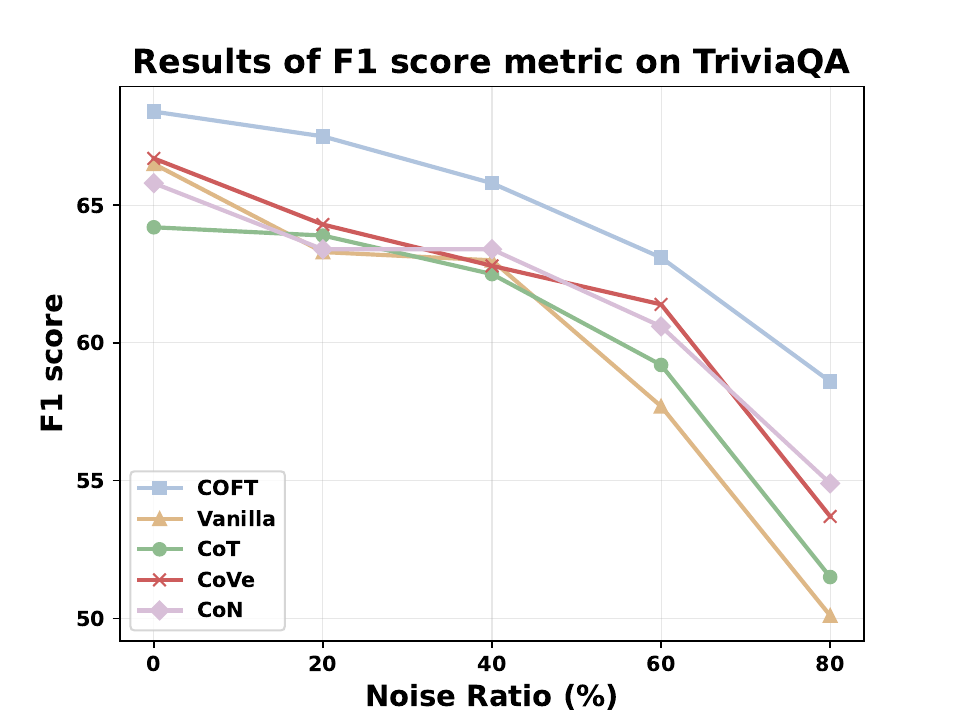}
        \includegraphics[width=0.32\textwidth, height=0.255\textwidth]{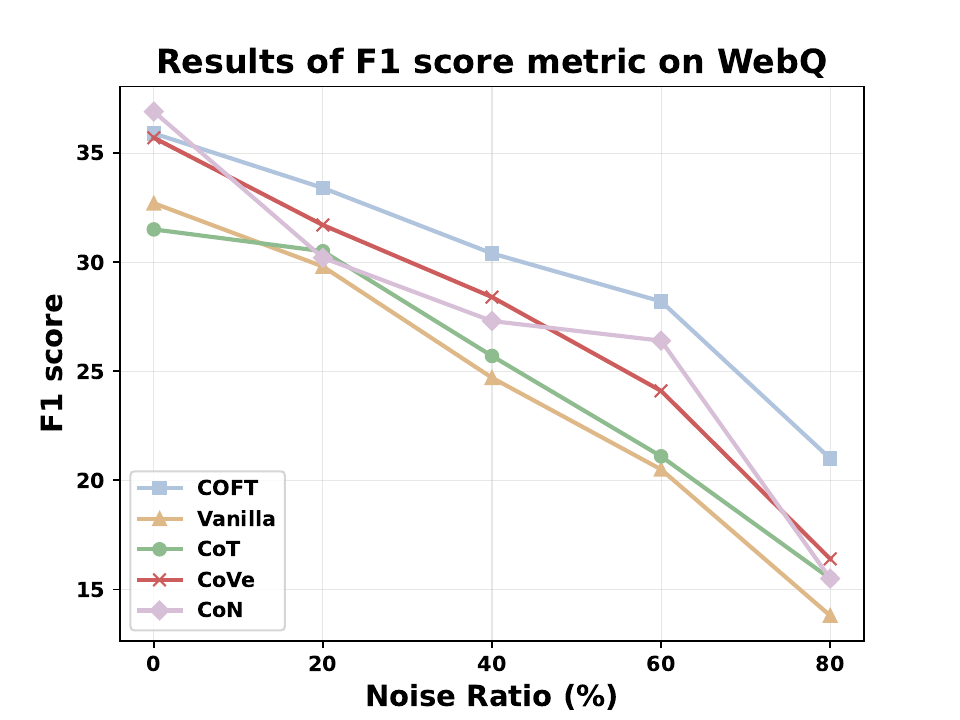}
        \caption{Evaluation on F1 score metric of noise robustness in question answering task, utilizing Vicuna-33B as the backbone model: COFT demonstrates superior performance on all three open-domain QA benchmarks, especially at higher noise ratios.}
        \label{fig:vicuna_noise_roubtness_f1}
    \end{minipage}
\end{figure*}

\begin{figure*}[t]
    \centering
    \begin{minipage}{1\textwidth}
        \includegraphics[width=0.32\textwidth, height=0.255\textwidth]{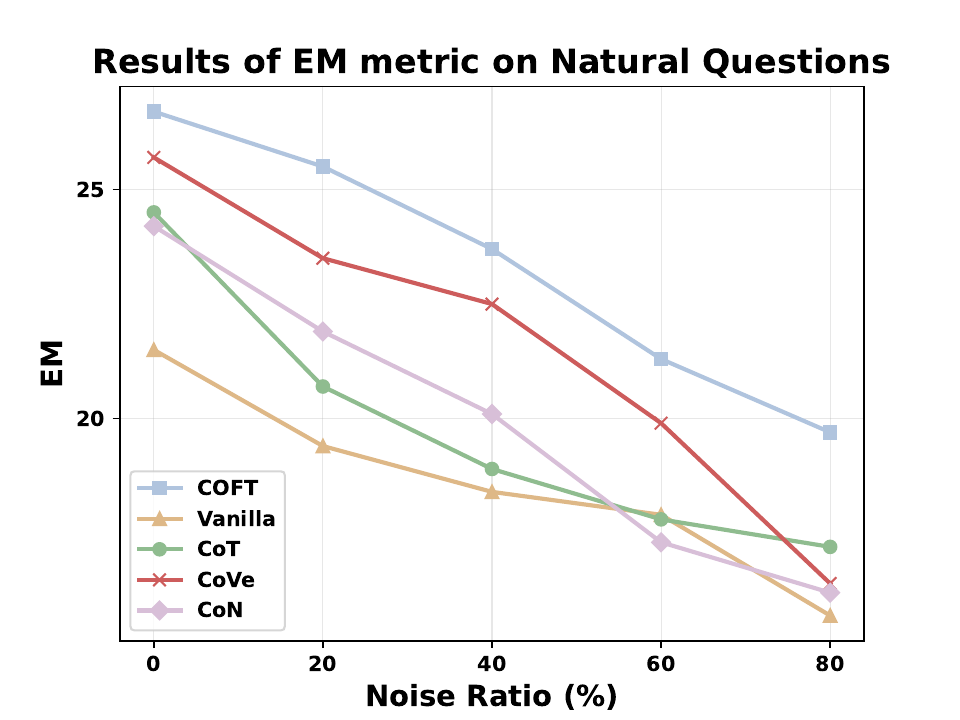}
        \includegraphics[width=0.32\textwidth, height=0.255\textwidth]{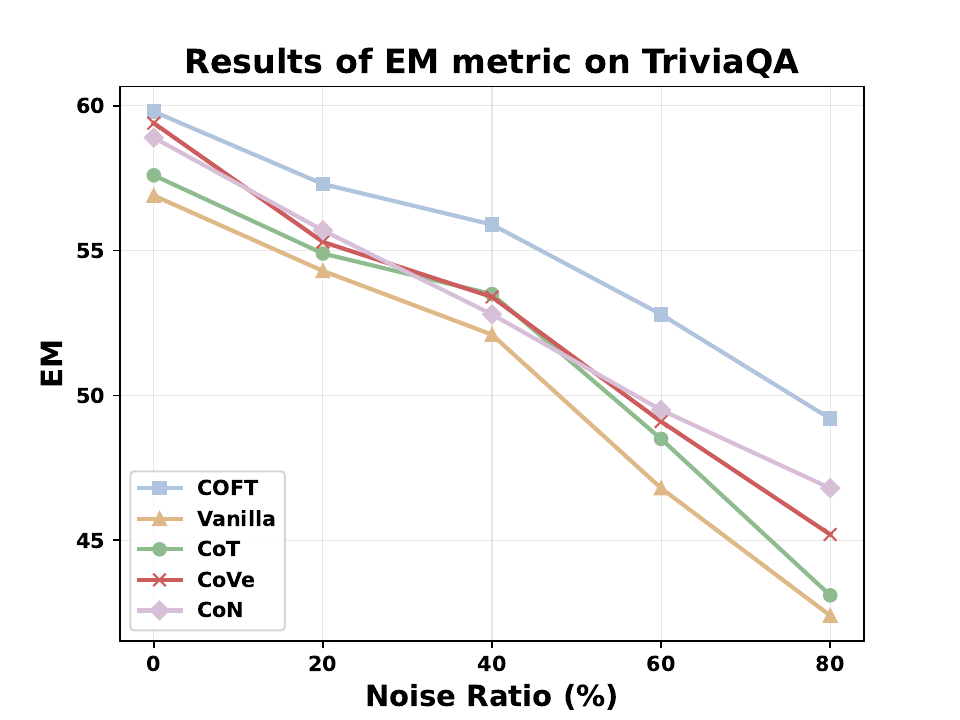}
        \includegraphics[width=0.32\textwidth, height=0.255\textwidth]{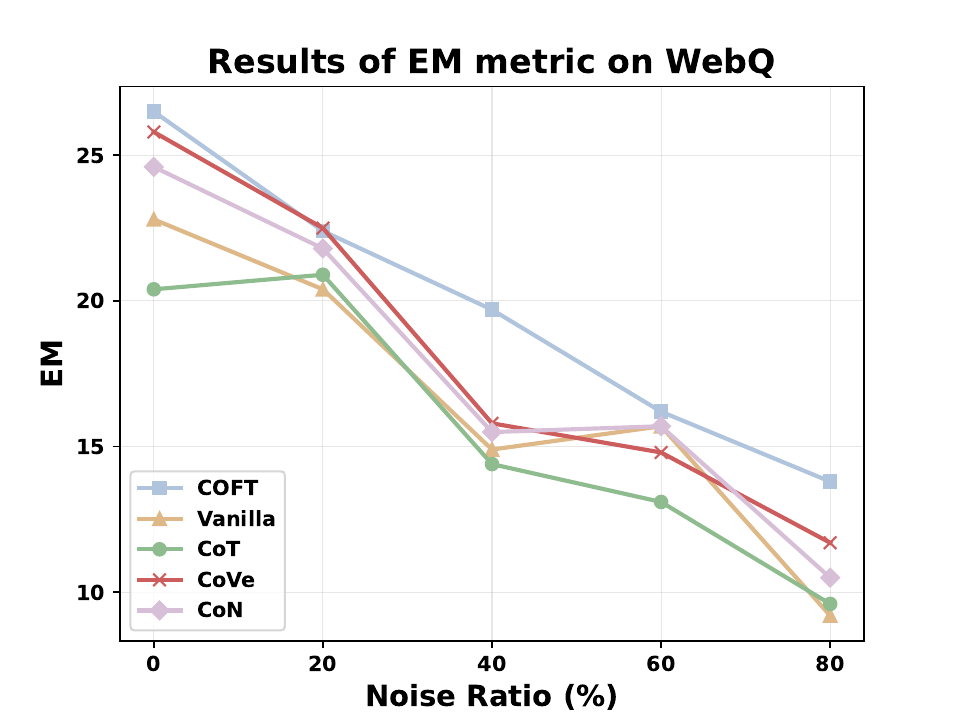}
        \caption{Evaluation on F1 score metric of noise robustness in question answering task, utilizing Vicuna-33B as the backbone model: COFT demonstrates superior performance on all three open-domain QA benchmarks, especially at higher noise ratios.}
        \label{fig:vicuna_noise_roubtness_EM}
    \end{minipage}
\end{figure*}

COFT also exhibits robustness against noise texts present in the reference contexts. We provide more question answering results illustrated in Figures \ref{fig:chatgpt_noise_roubtness_EM}, \ref{fig:GPT4_noise_roubtness_f1}, \ref{fig:GPT4_noise_roubtness_EM}, \ref{fig:vicuna_noise_roubtness_f1}, and \ref{fig:vicuna_noise_roubtness_EM}. We also provide detailed data tables as a numerical complement to the visual results in Section \ref{exp:qa}. As illustrated in Tables \ref{Tab:qa2}, \ref{Tab:qa1}, and \ref{Tab:qa3}, we observe that COFT is capable of maintaining relative robustness compared to other methods under conditions of severe noisy scenarios. 
This also demonstrates the effectiveness of mining key lexical and phrases relevant to the query. We further observe that as the noise ratio increases, that is, a greater proportion of irrelevant text in the reference context, COFT demonstrates enhanced robustness compared to other methods, thereby yielding relatively superior results. COFT achieves improvements or comparable results to baseline methods across nearly all conditions of \textit{noise ratio}. Notably,  under conditions where the \textit{noise ratio} is $80\%$, COFT achieves a maximum improvement of $6.5\%$ in EM metric and $10.5\%$ in the F1 score metric when utilizing ChatGPT as the backbone model, which also demonstrates the noise robustness under similar nosy documents and the capability to focus on 
the highlighted key lexical units to the given query of our COFT.

 Specifically, when ChatGPT serves as the backbone model and the \textit{noise ratio} goes from $0\%$ to $80\%$, on the Natural Questions dataset, COFT achieves average improvements of $4.3\%$ in the EM metric and $3.4\%$ in the F1 score metric. On the TriviaQA dataset, our COFT achieves average improvements of $2.7\%$ in the EM metric, alongside $2.0\%$ in the F1 score metric. Furthermore, on the WebQ dataset, COFT achieves average improvements of $2.3\%$ in the EM metric and $4.7\%$ in the F1 score metric. These results underscore the efficacy of the COFT approach in enhancing the performance of ChatGPT. 

When Vicuna-33B serves as the backbone model, COFT  has demonstrated notable improvements across different evaluation metrics. On the Natural Questions dataset, COFT achieves average improvements of $1.6\%$ in the EM metric and $3.0\%$ in the F1 score metric. On the TriviaQA dataset, COFT achieves average improvements of $2.0\%$ in the EM metric, along with $2.5\%$ in the F1 score metric. Furthermore, on the WebQ dataset, COFT achieves average improvements of $1.4\%$ in the EM metric and $1.8\%$ in the F1 score metric. These results underscore the efficacy of the COFT approach in enhancing the performance of Vicuna-33B.

When GPT4 serves as the backbone model, our  COFT exhibits enhancements across various evaluation metrics as well. Specifically, on the Natural Questions dataset, COFT achieves average improvements of $1.5\%$ in the EM metric and $2.0\%$ in the F1 score metric. On the TriviaQA dataset, COFT achieves average improvements of $3.4\%$ in the EM metric, along with $0.6\%$ in the F1 score metric. Moreover, on the WebQ dataset, COFT achieves average improvements of $4.6\%$ in the EM metric and $2.5\%$ in the F1 score metric. These results highlight the effectiveness of COFT in enhancing the performance of GPT4 in these question-answering tasks. 

These results further highlight COFT's efficacy in comparison to canonical methods. Such robustness to lengthy and noisy texts closely aligns with the real-world scenarios of user prompt inputs, effectively aiding LLMs in delivering more accurate responses. 
This effectively facilitates the practical deployment of LLMs in scenarios where high-precision and reliable answers are critically essential. Moreover, the effectiveness across various LLM backbones also demonstrates the potential of COFT to serve as a versatile plug-and-play framework over a wide range of long-form downstream NLP tasks.

\section{More Results of Ablation Study} \label{app:more_ab}

\subsection{Detailed Ablation Results of \textit{Selector}}\label{app:more_ab1}
In Section \ref{sec:ab}, we conduct the ablation study on \textit{selector} by setting the threshold to a fixed value of $0.5$, utilizing ChatGPT as the backbone model. In this section, we conduct a more detailed ablation study of the \textit{selector}. We experiment with the threshold 
$\tau$ for \textit{selector}, ranging from $0.1$ to $1.0$, and report the ablated results of Vicuna-33B, ChatGPT, and GPT4, respectively to provide more insight into our dynamic threshold algorithm. 

As shown in Tables \ref{Tab:thresholf_gpt}, \ref{Tab:thresholf_gpt4}, and \ref{Tab:thresholf_vicuna}. 
We still observe that our dynamic threshold algorithm achieves consistently superior and robust results against all other fixed thresholds. This effectively demonstrates the necessity of considering both the length and the amount of information of a given input reference context when setting the filtering thresholds to key lexical units. Moreover, the proposed dynamic threshold algorithm may potentially be beneficial to consider additional factors or optimize the combination method of context length and the amount of information to get improved results and we leave the exploration as a future work.

\subsection{Ablation Results for Vicuna-33B and GPT4} \label{app:more_ab2}
In Section \ref{sec:ab}, we report the results of the ablation study using ChatGPT as the backbone model. In this section, we will further present the results using Vicuna-33B and GPT4 as backbone models to obtain more insights into the individual components constituting COFT across various backbone models. As illustrated in Tables \ref{Tab:ab2} and \ref{Tab:ab3}, we still observe that the absence of each component within COFT invariably leads to a decline in performance across diverse domains for Vicuna-33B and GPT4 in the FELM benchmark, which demonstrates that COFT organically integrates these components into a unified framework as well. 

Remarkably, we observe that in the absence of a \textit{scorer}, i.e., \textit{selector} randomly retains the top $\tau \times 100\%$ of key candidates obtained by the \textit{recaller} using a dynamic threshold algorithm, rather than preserving them in descending order based on contextual weight, leads to a more significant decline in performance. This underscores the critical importance of effectively measuring the candidates' significance and highlights these candidates in reducing the issue of knowledge hallucination within LLMs as well.


These results underscore the organic integration of the three core components of COFT, \textit{recaller}, \textit{scorer}, and \textit{selector}. This integration is not merely additive but forms a cohesive framework that significantly reduces the problem of knowledge hallucination in LLMs.

\begin{table*}[htbp]
	\caption{A more detailed ablation study for \textit{selector} on the knowledge hallucination benchmark, FELM, using ChatGPT as the backbone model. To demonstrate the superiority of our dynamic threshold algorithm, we set the threshold $\tau$ ranging from $0.1$ to $1.0$.}\label{Tab:thresholf_gpt}
  \vspace{2mm}
	\centering
	\resizebox{\columnwidth}{!}{
		\begin{tabular}{l l c c c c c c c c c} 
			\toprule
			& & \multicolumn{3}{c}{\textbf
   {WK}} & \multicolumn{3}{c}{\textbf{Sci/Tech}} & \multicolumn{3}{c}{\textbf{Wri/Rec}} \\
			\cmidrule(lr){3-5} \cmidrule(lr){6-8} \cmidrule(lr){9-11}
			Backbone & Threshold & F1 Score & Precision & Recall & F1 Score & Precision & Recall & F1 Score & Precision & Recall \\
			\midrule
			 \multirow{11}{*}{ChatGPT} & $\tau=0.1$ & 76.8 & 77.9& 75.7 & 78.0 & 82.6 & 74.0 & 71.8 & 87.5 & 60.9 \\
    			 & $\tau=0.2$ & 69.8 & 70.6 & 69.1 & 72.6 & 78.2 & 67.7 & 68.1  & 85.1 & 56.7 \\
        			 & $\tau=0.3$ & 77.5 & 82.0 & 73.5 & 74.5 & 80.9 & 69.0 & 73.9 & 87.3 & 64.1 \\
			 & $\tau=0.4$ & 75.2 & 82.5& 69.1 & 74.5 & 84.3 & 66.7 & 71.5 & 87.3 & 60.5 \\
			 & $\tau=0.5$ & 77.3 & 79.7 & 75.1 & 75.7 & 82.4 & 70.1 & 70.7 & 86.1 & 60.1\\
    	& $\tau=0.6$ & 78.7 & 84.7& 63.2 & 81.4 & 84.8 & 78.3 & 72.3 & 85.1 & 62.8 \\
        & $\tau=0.7$ & 75.4 & 75.1& 75.7 & 81.6 & 81.3 & 81.9 & 56.6 & 81.8 & 43.3 \\
            & $\tau=0.8$ & 66.2 & 69.4& 63.2 & 72.4 & 80.4 & 65.9 & 70.8 & 84.4 & 60.9 \\
            & $\tau=0.9$  & 74.5 & 73.9& 75.0 & 61.7 & 86.1 & 48.1 & 69.3 & 88.0 & 57.1 \\
            & $\tau=1.0$ & 77.9 & 81.5& 74.6 & 82.8 & 83.4 & 82.2 & 61.2 & 82.2 & 48.7 \\
			\cmidrule{2-11}
			& COFT & \textbf{81.6} & \textbf{85.5} & \textbf{77.9} & \textbf{85.5} & \textbf{86.5} & \textbf{84.5} & \textbf{75.2 }& \textbf{88.3} & \textbf{65.4} \\ 
			\bottomrule
   
		\end{tabular}
	}
 \end{table*}

\begin{table*}[htbp]
	\caption{A more detailed ablation study for \textit{selector} on the knowledge hallucination benchmark, FELM, using Vicuna-33B as the backbone model. To demonstrate the superiority of our dynamic threshold algorithm, we set the threshold $\tau$ ranging from $0.1$ to $1.0$.}\label{Tab:thresholf_gpt4}
 \vspace{2mm}
	\centering
	\resizebox{\columnwidth}{!}{
		\begin{tabular}{l l c c c c c c c c c} 
			\toprule
			& & \multicolumn{3}{c}{\textbf
   {WK}} & \multicolumn{3}{c}{\textbf{Sci/Tech}} & \multicolumn{3}{c}{\textbf{Wri/Rec}} \\
			\cmidrule(lr){3-5} \cmidrule(lr){6-8} \cmidrule(lr){9-11}
			Backbone & Threshold & F1 Score & Precision & Recall & F1 Score & Precision & Recall & F1 Score & Precision & Recall \\
			\midrule
			 \multirow{11}{*}{Vicuna-33B} & $\tau=0.1$ & 56.6 & 55.8& 57.5 & 63.2 & 64.0 & 62.4 & 60.8 & 52.3 & 72.5 \\
    			 & $\tau=0.2$ & 54.2 & 53.9& 55.1 & 65.9 & 53.4 & 70.2 & 72.3 & 62.1 & 86.4 \\
        			 & $\tau=0.3$ & 54.0 & 56.8& 51.5 & 67.3 & 63.4 & 71.7 & 69.7 & 62.8 & 78.2 \\
			 & $\tau=0.4$ & 52.9 & 54.5& 51.3 & 69.7 & 62.4 & 75.2 &67.5  & 56.8 & 83.3 \\
			 & $\tau=0.5$ & 57.3 & 53.7 & 61.5 & 52.6 &50.8 & 54.5 &60.1 & 59.3&61.1\\
    	& $\tau=0.6$ & 57.5 & 57.5 & 57.6 & 67.0 & 62.7 & 72.0 & 73.2 & 64.7 & 84.2 \\
        & $\tau=0.7$ & 60.9 & 59.6& 62.2 & 67.6 & 60.6 & 76.5 & 67.8 & 60.3 & 77.4 \\
            & $\tau=0.8$ & 53.0 & 52.5& 53.6 & 66.1 & 58.7 & 75.6 & 62.6 & 51.9 & 78.8 \\
            & $\tau=0.9$ & 50.2 & 64.1& 41.3 & 63.8 & 62.9 & 64.7 & 65.4 & 60.3 & 71.5 \\
            & $\tau=1.0$ & 52.2 & 48.9& 55.9 & 62.4 & 59.1 & 66.1 & 66.5 & 55.5 & 82.8 \\
			\cmidrule{2-11}
         & COFT & \textbf{64.4} & \textbf{61.7} & \textbf{67.4} & \textbf{70.9} & \textbf{65.7} & \textbf{77.2} & \textbf{77.3} & \textbf{67.9} & \textbf{89.8} \\
			\bottomrule
   
		\end{tabular}
	}
 \end{table*}

\begin{table*}[htbp]
	\caption{A more detailed ablation study for \textit{selector} on the knowledge hallucination benchmark, FELM, using GPT4 as the backbone model. To demonstrate the superiority of our dynamic threshold algorithm, we set the threshold $\tau$ ranging from $0.1$ to $1.0$.}\label{Tab:thresholf_vicuna}
  \vspace{2mm}
	\centering
	\resizebox{\columnwidth}{!}{
		\begin{tabular}{l l c c c c c c c c c} 
			\toprule
			& & \multicolumn{3}{c}{\textbf
   {WK}} & \multicolumn{3}{c}{\textbf{Sci/Tech}} & \multicolumn{3}{c}{\textbf{Wri/Rec}} \\
			\cmidrule(lr){3-5} \cmidrule(lr){6-8} \cmidrule(lr){9-11}
			Backbone & Threshold & F1 Score & Precision & Recall & F1 Score & Precision & Recall & F1 Score & Precision & Recall \\
			\midrule
			 \multirow{11}{*}{GPT4} & $\tau=0.1$ & 74.6 & 89.3& 64.1 & 65.4 & 80.7 & 55.0 & 61.0 & 78.3 & 50.0 \\
    			 & $\tau=0.2$ & 73.7 & 92.0& 61.5 &69.1 &  80.7 & 55.0 & 67.8 & 62.1 & 74.7 \\
        			 & $\tau=0.3$ & 67.7 & 91.3& 53.8 & 72.7 & 82.4 & 65.1 & 44.2 & 63.0 & 34.0 \\
			 & $\tau=0.4$ & 75.8 & 92.6& 64.1 & 65.1 & 81.4 & 54.3 & 69.5 & 63.0 & 77.5 \\
			 & $\tau=0.5$ & 77.9 &83.4 & 73.1 & 73.4 &  76.5 & 70.6&79.4 &83.5 & 75.6\\
    	& $\tau=0.6$ & 83.3 & 90.9& 76.9 & 48.1 & 75.4 & 35.3 & 78.5 & 84.3 & 73.5 \\
        & $\tau=0.7$ & 74.6 & 89.3& 64.1 & 68.7 & 82.5 & 58.8 & 80.6 & 84.7 & 76.9 \\
            & $\tau=0.8$ & 76.5 & 89.7& 66.7 & 67.8 & 77.9 & 60.0 &77.9 & 79.2 & 76.7 \\
            & $\tau=0.9$ & 80.0 & 90.3& 71.8 & 61.5 & 79.9 & 50.0 & 67.3 & 78.3 & 59.0 \\
            & $\tau=1.0$ & 80.1 & 90.3& 71.8 & 36.3 & 84.5 & 23.1 & 57.1 & 70.6 & 48.0 \\
			\cmidrule{2-11}
			& COFT  & \textbf{87.3} & \textbf{94.8} &  \textbf{{80.9}} & \textbf{{77.9}} & \textbf{86.0} & \textbf{{71.3}} & \textbf{{84.5}} & \textbf{92.9} & \textbf{{77.9}} \\  
			\bottomrule
   
		\end{tabular}
	}
 \end{table*}



\section{Inference Time Comparisons}
We note that COFT requires an additional process of highlighting the input text before feeding it into the LLM for reasoning. Compared to the vanilla model, this process could potentially introduce extra inference time. Hence, in this section, we record and compare the average inference time per sample of different methods including vanilla, CoT, RALM, CoN, CoVe, and COFT on the knowledge hallucination benchmark, FELM, to explore the influence of additional inference time and provide more insight of our COFT. We report the word-level granularity COFT as an example, as the inference times for COFT at three different granularity levels (paragraph level, sentence level, and word level) are nearly identical.

We report the average inference time per sample as a metric, as shown in Table \ref{tab:inference_time}.
We observe from the table that although the incorporation of COFT as a preprocessing module for LLM introduces additional inference time costs, this impact is marginal. On average, the increase in inference time cost per sample due to the introduction of COFT, compared to the vanilla model, is $12\%$. Notably, when utilizing accelerated APIs such as GPT, this additional inference time is less than one second, yet it offers an average improvement of 33.2\% and a maximum of 60.5\% in the F1 score for existing LLMs to reduce the issue of knowledge hallucination. Furthermore, COFT exhibits higher inference efficiency compared to methods such as CoN, and CoVe, indicating that the additional computational overhead introduced by COFT is limited. We may focus on exploring ways to further reduce the time cost of the COFT, including lightweighting small language models to get a faster calculation of contextual weight\cite{qinglianghua,dist1} or adopting more efficient \cite{bitnet, qmoe, atom} and rational large model inference strategies such as speculative decoding \cite{tuili1, tuili2} as for future works.

\begin{figure*}[t]
    \centering 
    \includegraphics[width=\columnwidth]{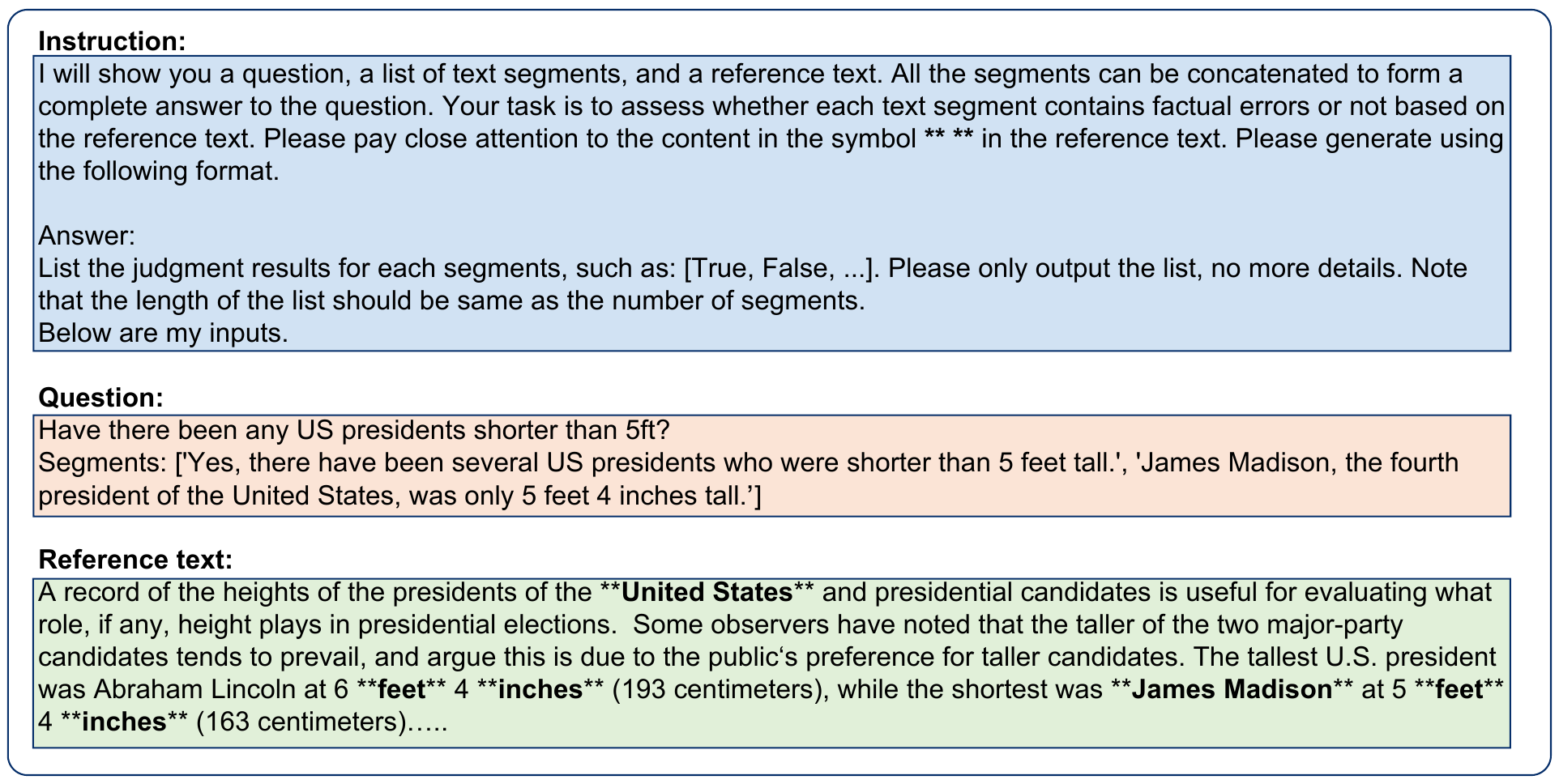}
    \caption{Prompt templates of knowledge hallucination task after highlighting the key lexical units. We use a sample prompt template across Vicuna-33B, ChatGPT, and GPT4.}
    \label{case_felm}
\end{figure*}

\begin{figure*}[t]
    \centering 
    \includegraphics[width=\columnwidth]{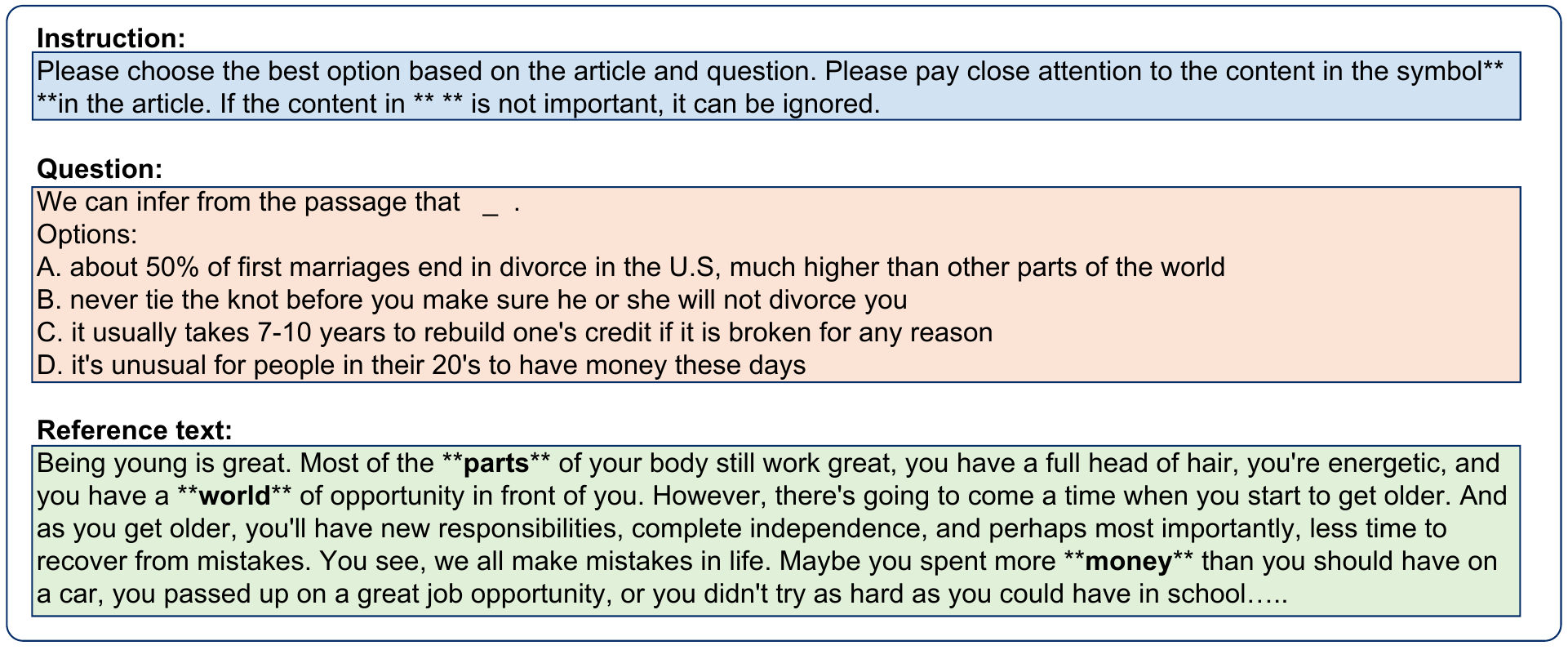}
    \caption{Prompt templates of reading comprehension task after highlighting the key lexical units. We use a sample prompt template across Vicuna-33B, ChatGPT, and GPT4.}
    \label{case_rc}
\end{figure*}

\begin{figure*}[t]
    \centering 
    \includegraphics[width=\columnwidth]{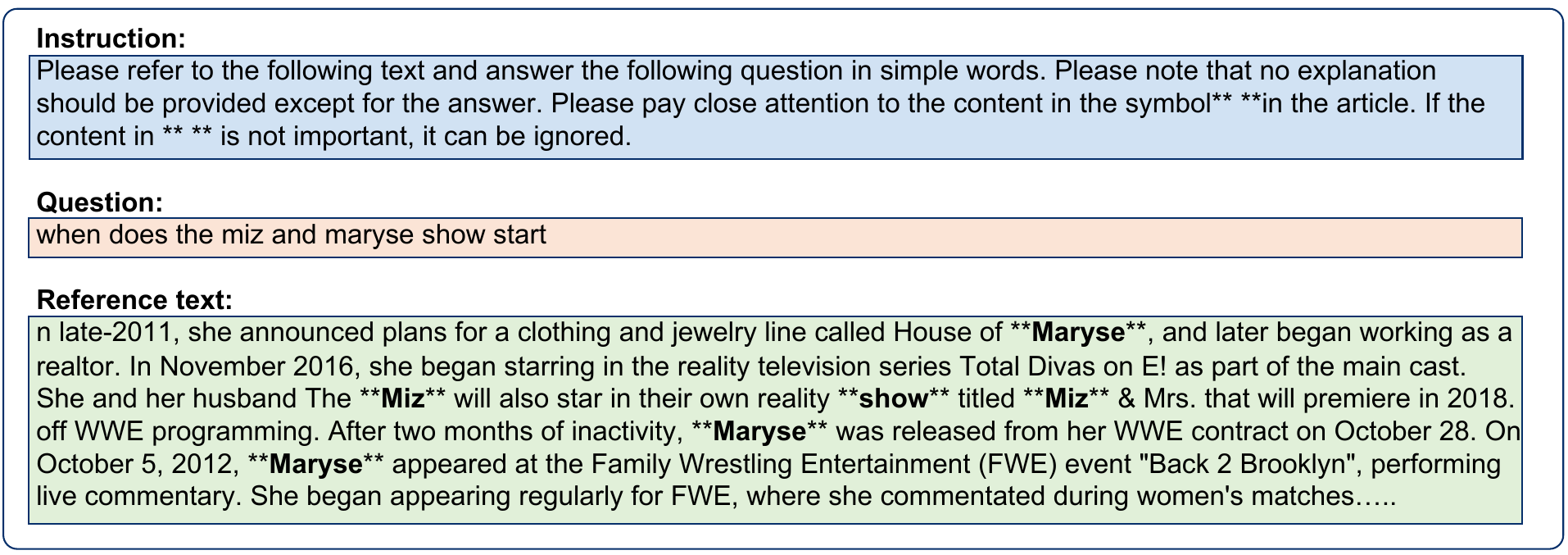}
    \caption{Prompt templates of question answering task after highlighting the key lexical units. We use a sample prompt template across Vicuna-33B, ChatGPT, and GPT4.}
    \label{case_nq}
\end{figure*}

\section{More Results of Smaller Self-information Calculator}

\begin{table*}[ht]
	\caption{Results of knowledge hallucination benchmark on WK (world knowledge), Sci/Tech (science and technology), and Wri/Rec (writing and recommendation) domains using GPT2-small (124M) to calculate the contextual weight. We denote COFT at the paragraph, sentence, and word levels as COFT\(_{p}\), COFT\(_{s}\), and COFT\(_{w}\),  respectively.}\label{Tab:hallu_gtp_small}
	\centering
	\resizebox{\columnwidth}{!}{%
		\begin{tabular}{l l c c c c c c c c c} 
			\toprule
			Backbone & Method & \multicolumn{3}{c}{WK} & \multicolumn{3}{c}{Sci/Tech} & \multicolumn{3}{c}{Wri/Rec} \\
			\cmidrule(lr){3-5} \cmidrule(lr){6-8} \cmidrule(lr){9-11}
			 &  & F1 Score & Precision & Recall & F1 Score & Precision & Recall & F1 Score & Precision & Recall \\
			\midrule
			\multirow{3}{*}{Vicuna-33B} 
			& \(COFT_w\) & 63.7 & 62.4 & 65.1 & 69.1 & 63.8 & 75.3 & 74.6 & 69.8 & 80.1 \\
			& \(COFT_s\) & 61.3 & 60.6 & 62.1 & 66.9 & 63.9 & 70.1 & 69.9 & 65.5 & 74.9 \\
			& \(COFT_p\) & 64.2 & 65.3 & 63.2 & 69.2 & 66.6 & 72.1 & 75.9 & 69.9 & 83.1 \\
			\midrule
			\multirow{3}{*}{ChatGPT} 
			& \(COFT_w\) & 74.0 & 77.9 & 70.4 & 79.6 & 79.1 & 80.1 & 78.1 & 86.1 & 71.5 \\
			& \(COFT_s\) & 71.8 & 70.5 & 73.1 & 74.3 & 75.5 & 73.1 & 75.6 & 77.9 & 73.5 \\
			& \(COFT_p\) & 77.3 & 81.8 & 73.3 & 79.6 & 75.3 & 84.4 & 79.7 & 85.5 & 74.7 \\
			\midrule
			\multirow{3}{*}{GPT-4} 
			& \(COFT_w\) & 74.0 & 77.9 & 70.4 & 79.6 & 79.1 & 80.1 & 78.1 & 86.1 & 71.5 \\
			& \(COFT_s\) & 71.8 & 70.5 & 73.1 & 74.3 & 75.5 & 73.1 & 75.6 & 77.9 & 73.5 \\
			& \(COFT_p\) & 77.3 & 81.8 & 73.3 & 79.6 & 75.3 & 84.4 & 79.7 & 85.5 & 74.7 \\
			\bottomrule
		\end{tabular}
	}
\end{table*}

\begin{table*}[ht]
	\caption{Results of knowledge hallucination benchmark on WK (world knowledge), Sci/Tech (science and technology), and Wri/Rec (writing and recommendation) domains using GPT2-medium (355M) to calculate the contextual weight. We denote COFT at the paragraph, sentence, and word levels as COFT\(_{p}\), COFT\(_{s}\), and COFT\(_{w}\), respectively.}\label{Tab:hallu_gtp_medium}
	\centering
	\resizebox{\textwidth}{!}{%
		\begin{tabular}{l l c c c c c c c c c} 
			\toprule
			Backbone & Method & \multicolumn{3}{c}{WK} & \multicolumn{3}{c}{Sci/Tech} & \multicolumn{3}{c}{Wri/Rec} \\
			\cmidrule(lr){3-5} \cmidrule(lr){6-8} \cmidrule(lr){9-11}
			 &  & F1 Score & Precision & Recall & F1 Score & Precision & Recall & F1 Score & Precision & Recall \\
			\midrule
			\multirow{3}{*}{Vicuna-33B} 
			& COFT\_w & 65.2 & 62.0 & 68.8 & 70.8 & 66.3 & 76.0 & 72.6 & 62.8 & 85.9 \\
			& COFT\_s & 62.8 & 60.6 & 65.2 & 67.9 & 63.1 & 73.5 & 69.5 & 65.5 & 74.0 \\
			& COFT\_p & 66.9 & 66.8 & 67.1 & 65.0 & 60.6 & 70.1 & 71.2 & 62.7 & 82.4 \\
			\midrule
			\multirow{3}{*}{ChatGPT} 
			& COFT\_w & 75.8 & 80.1 & 71.9 & 72.4 & 85.3 & 62.9 & 77.5 & 79.9 & 75.3 \\
			& COFT\_s & 70.9 & 77.9 & 65.1 & 73.7 & 80.7 & 67.9 & 74.8 & 85.1 & 66.8 \\
			& COFT\_p & 80.4 & 83.5 & 77.5 & 77.4 & 83.5 & 72.1 & 78.7 & 84.6 & 73.6 \\
			\midrule
			\multirow{3}{*}{GPT-4} 
			& COFT\_w & 81.1 & 83.3 & 79.1 & 79.7 & 80.1 & 79.4 & 84.5 & 80.1 & 89.3 \\
			& COFT\_s & 79.1 & 85.1 & 73.9 & 76.9 & 80.3 & 73.7 & 80.8 & 83.2 & 78.5 \\
			& COFT\_p & 81.8 & 85.9 & 78.0 & 76.4 & 79.1 & 73.9 & 80.9 & 90.5 & 73.1 \\
			\bottomrule
		\end{tabular}
	}
\end{table*}

\begin{table*}[ht]
    \caption{Results of knowledge hallucination benchmark on WK (world knowledge), Sci/Tech (science and technology), and Wri/Rec (writing and recommendation) domains using GPT2-large (744M) to calculate the contextual weight. We denote COFT at the paragraph, sentence, and word levels as COFT\(_{p}\), COFT\(_{s}\), and COFT\(_{w}\), respectively.}\label{Tab:hallu_gtp_large}
    \centering
    \resizebox{\textwidth}{!}{%
        \begin{tabular}{l l c c c c c c c c c} 
            \toprule
            Backbone & Method & \multicolumn{3}{c}{WK} & \multicolumn{3}{c}{Sci/Tech} & \multicolumn{3}{c}{Wri/Rec} \\
            \cmidrule(lr){3-5} \cmidrule(lr){6-8} \cmidrule(lr){9-11}
             &  & F1 Score & Precision & Recall & F1 Score & Precision & Recall & F1 Score & Precision & Recall \\
            \midrule
            \multirow{3}{*}{Vicuna-33B} 
            & COFT\_w & 63.3 & 61.1 & 65.7 & 69.8 & 63.2 & 78.0 & 75.8 & 65.9 & 89.3 \\
            & COFT\_s & 63.2 & 62.6 & 63.8 & 69.3 & 64.5 & 74.8 & 72.8 & 60.6 & 91.2 \\
            & COFT\_p & 63.6 & 60.4 & 67.2 & 67.0 & 61.3 & 73.9 & 70.3 & 64.8 & 76.9 \\
            \midrule
            \multirow{3}{*}{ChatGPT} 
            & COFT\_w & 73.7 & 76.1 & 71.5 & 80.2 & 73.5 & 88.2 & 78.4 & 79.3 & 77.6 \\
            & COFT\_s & 75.2 & 69.9 & 81.3 & 74.1 & 83.1 & 66.9 & 73.9 & 74.6 & 73.3 \\
            & COFT\_p & 75.9 & 74.5 & 77.4 & 83.1 & 82.5 & 83.7 & 77.3 & 80.6 & 74.2 \\
            \midrule
            \multirow{3}{*}{GPT-4} 
            & COFT\_w & 82.0 & 85.2 & 79.1 & 84.3 & 80.1 & 88.9 & 88.3 & 84.1 & 92.9 \\
            & COFT\_s & 78.2 & 88.5 & 70.1 & 75.8 & 78.4 & 73.3 & 86.3 & 85.3 & 87.4 \\
            & COFT\_p & 82.9 & 87.9 & 78.4 & 80.6 & 82.8 & 78.5 & 86.5 & 88.2 & 84.9 \\
            \bottomrule
        \end{tabular}
    }
\end{table*}

\begin{table*}[ht]
	\caption{Results of knowledge hallucination benchmark on WK (world knowledge), Sci/Tech (science and technology), and Wri/Rec (writing and recommendation) domains using GPT2-XL (1.5B) to calculate the contextual weight. We denote COFT at the paragraph, sentence, and word levels as COFT\(_{p}\), COFT\(_{s}\), and COFT\(_{w}\), respectively.}\label{Tab:hallu_gtp_xl}
	\centering
	\resizebox{\columnwidth}{!}{%
		\begin{tabular}{l l c c c c c c c c c} 
			\toprule
			Backbone & Method & \multicolumn{3}{c}{WK} & \multicolumn{3}{c}{Sci/Tech} & \multicolumn{3}{c}{Wri/Rec} \\
			\cmidrule(lr){3-5} \cmidrule(lr){6-8} \cmidrule(lr){9-11}
			 &  & F1 Score & Precision & Recall & F1 Score & Precision & Recall & F1 Score & Precision & Recall \\
			\midrule
			\multirow{3}{*}{Vicuna-33B} 
			& \(COFT_w\) & 63.5 & 62.9 & 64.1 & 69.5 & 65.5 & 74.1 & 78.6 & 71.3 & 87.5 \\
			& \(COFT_s\) & 66.0 & 73.1 & 60.1 & 65.2 & 71.3 & 60.1 & 74.5 & 66.9 & 84.1 \\
			& \(COFT_p\) & 65.2 & 69.1 & 61.7 & 69.4 & 67.2 & 71.8 & 71.5 & 65.1 & 79.3 \\
			\midrule
			\multirow{3}{*}{ChatGPT} 
			& \(COFT_w\) & 73.5 & 81.5 & 66.9 & 79.4 & 74.2 & 85.4 & 74.7 & 86.4 & 65.8 \\
			& \(COFT_s\) & 76.1 & 83.3 & 70.1 & 78.1 & 73.9 & 82.7 & 76.7 & 80.5 & 73.3 \\
			& \(COFT_p\) & 77.9 & 80.5 & 75.4 & 81.3 & 77.8 & 85.2 & 78.7 & 84.1 & 73.9 \\
			\midrule
			\multirow{3}{*}{GPT4} 
			& \(COFT_w\) & 82.1 & 83.8 & 80.5 & 83.8 & 79.2 & 89.0 & 88.1 & 88.3 & 87.9 \\
			& \(COFT_s\) & 80.1 & 82.9 & 77.5 & 80.1 & 82.9 & 77.4 & 82.7 & 79.6 & 86.1 \\
			& \(COFT_p\) & 85.0 & 89.4 & 81.1 & 82.7 & 79.9 & 85.7 & 85.0 & 84.7 & 85.3 \\
			\bottomrule
		\end{tabular}
	}
\end{table*}

In Section \ref{sec:hallu}, we utilize Llama 7B as the self-information calculator due to its great performance across a wide range of downstream tasks \cite{llama}. To further demonstrate the generalization and versatility of COFT across models of smaller scales, we conduct additional experiments using GPT-2 small (124M), GPT-2 medium (355M), GPT-2 large (744M), and GPT-2 XL (1.5B) to calculate the self-information, respectively \cite{gpt2}.  As shown in Tables \ref{Tab:hallu_gtp_small}, \ref{Tab:hallu_gtp_medium}, \ref{Tab:hallu_gtp_large} and \ref{Tab:hallu_gtp_xl}, COFT consistently exhibits superior performance across all baseline methods, which demonstrates the effectiveness and potentially broad applications to smaller models.

\section{More In-depth Analysis of COFT}

\subsection{Comparison Results of Adding Special Prompt}

 We conduct experiments of the baseline methods with the addition of the prompt "Please pay close attention to the most relevant content in the text" on the FELM benchmark for the knowledge hallucination task. As shown in Table \ref{tab:hallu_add_prompt} , we find that while the inclusion of this prompt leads to marginal improvements in the overall performance, it can also result in a decline in performance in certain cases. This suggests that simply appending the prompt can not consistently enhance performance.

In contrast, COFT has demonstrated significant improvements, with an average increase of 30\% and a maximum improvement of 60.5\% in F1 scores across the knowledge hallucination benchmark, which further demonstrates the great benefits of COFT.

\begin{table}[ht]
\centering
\caption{Results of knowledge hallucination benchmark on WK (world knowledge), Sci/Tech (science and technology), and Wri/Rec (writing and recommendation) domains by adding the prompt "Please pay close attention to the most relative content in the text".}
\label{tab:hallu_add_prompt}

\resizebox{\textwidth}{!}{
\begin{tabular}{@{}llccccccccccc@{}}
\toprule
& & \multicolumn{3}{c}{\textbf{WK}} & \multicolumn{3}{c}{\textbf{Sci/Tech}} & \multicolumn{3}{c}{\textbf{Wri/Rec}} \\
\cmidrule(lr){3-5} \cmidrule(lr){6-8} \cmidrule(lr){9-11}
\textbf{Backbone} & \textbf{Methods} & \textbf{F1 Score} & \textbf{Precision} & \textbf{Recall} & \textbf{F1 Score} & \textbf{Precision} & \textbf{Recall} & \textbf{F1 Score} & \textbf{Precision} & \textbf{Recall} \\
\midrule
\multirow{5}{*}{Vicuna-33B} & Vanilla & 35.1 & 29.6 & 43.1 & 26.2 & 17.9 & 48.8 & 30.3 & 19.2 & 72.0 \\
 & CoT & 35.0 & 31.1 & 40.1 & 24.5 & 15.9 & 53.3 & 30.5 & 19.9 & 65.4 \\
 & RALM & 47.2 & 44.2 & 50.6 & 36.1 & 27.3 & 53.1 & 31.2 & 19.4 & 80.1 \\
 & CoVe & 46.7 & 45.1 & 48.5 & 47.7 & 41.1 & 56.7 & 64.4 & 59.2 & 70.5 \\
 & CoN & 54.3 & 55.1 & 53.5 & 59.7 & 56.0 & 63.9 & 66.2 & 60.3 & 73.3 \\
\midrule
\multirow{5}{*}{ChatGPT} & Vanilla & 12.8 & 27.9 & 8.3 & 4.3 & 6.3 & 3.3 & 2.1 & 5.5 & 1.3 \\
 & CoT & 8.3 & 30.9 & 4.8 & 7.9 & 23.9 & 4.7 & 6.3 & 11.9 & 4.3 \\
 & RALM & 25.7 & 33.7 & 20.8 & 18.4 & 18.8 & 18.0 & 23.1 & 58.3 & 14.4 \\
 & CoVe & 18.8 & 46.7 & 11.8 & 18.8 & 13.7 & 29.8 & 21.1 & 60.4 & 12.8 \\
 & CoN & 18.0 & 63.0 & 10.5 & 21.9 & 24.4 & 19.9 & 31.3 & 28.9 & 34.2 \\
\midrule
\multirow{5}{*}{GPT-4} & Vanilla & 39.7 & 80.3 & 26.4 & 21.7 & 63.5 & 13.1 & 25.6 & 84.4 & 15.1 \\
 & CoT & 53.2 & 82.1 & 39.3 & 27.6 & 61.0 & 17.8 & 27.0 & 84.5 & 16.1 \\
 & RALM & 53.9 & 78.8 & 41.0 & 31.4 & 55.7 & 21.9 & 48.4 & 70.3 & 36.9 \\
 & CoVe & 49.2 & 50.5 & 47.9 & 70.1 & 85.5 & 59.4 & 50.9 & 51.2 & 50.7 \\
 & CoN & 55.0 & 50.1 & 61.0 & 68.7 & 80.3 & 60.1 & 69.3 & 75.1 & 64.3 \\
\bottomrule
\end{tabular}
}

\end{table}

\subsection{Two-hop Neighborhood Results of COFT}

COFT exhibits excellent scalability and can be extended to multi-hop neighbor situations. COFT initially focuses on one-hop neighbors of candidate entities within the KG due to their intrinsic relevance and close association. And leveraging only a single hop from the neighbors results in an average increase of 30\% and a maximum improvement of 60.5\% in the F1 score on the knowledge hallucination task.

We conduct additional experiments incorporating two-hop neighbor information of COFT on the knowledge hallucination benchmark. As shown in Table \ref{tab:hallu_two_hop}, compared to the one-hop version of COFT, integrating two-hop neighbor information further enriches the input provided to the LLMs, leading to a moderate performance improvement over the one-hop scenario.

Despite the primary focus on one-hop neighbors, COFT maintains great performance. This demonstrates the effectiveness of extracting one-hop neighbors. When incorporating two-hop information, COFT further achieves a better result, which also demonstrates the flexibility and scalability of COFT. Therefore, for more complex question scenarios, there are also potential benefits of incorporating two-hop or even multi-hop neighbors to further increase the performance of COFT.

\begin{table}[ht]
	\centering
	\caption{Results of knowledge hallucination benchmark on WK (world knowledge), Sci/Tech (science and technology), and Wri/Rec (writing and recommendation) domains by incorporating two-hop neighbor information of COFT. We denote COFT at the paragraph, sentence, and word levels by COFT\_p, COFT\_s, and COFT\_w, respectively.}
	\label{tab:hallu_two_hop}
	\resizebox{\textwidth}{!}{%
	\begin{tabular}{llccccccccccc}
		\toprule
		& & \multicolumn{3}{c}{WK} & \multicolumn{3}{c}{Sci/Tech} & \multicolumn{3}{c}{Wri/Rec} \\
		\cmidrule(lr){3-5} \cmidrule(lr){6-8} \cmidrule(lr){9-11}
		Backbone & Methods & F1 Score & Precision & Recall & F1 Score & Precision & Recall & F1 Score & Precision & Recall \\
		\midrule
		\multirow{3}{*}{Vicuna-33B} & COFT\_p & 69.5 & 73.3 & 66.0 & 67.0 & 63.5 & 71.0 & 69.2 & 64.3 & 75.0 \\
		 & COFT\_s & 61.9 & 60.6 & 63.2 & 72.1 & 69.5 & 74.9 & 67.2 & 61.5 & 74.1 \\
		 & COFT\_w & 63.7 & 63.5 & 64.0 & 68.3 & 63.3 & 74.2 & 77.0 & 74.8 & 79.4 \\
		\midrule
		\multirow{3}{*}{ChatGPT} & COFT\_p & 78.4 & 79.5 & 77.4 & 83.5 & 83.1 & 83.9 & 78.7 & 90.5 & 69.7 \\
		 & COFT\_s & 76.1 & 77.1 & 75.1 & 78.3 & 75.5 & 81.3 & 79.2 & 94.9 & 67.9 \\
		 & COFT\_w & 82.2 & 86.8 & 78.1 & 86.4 & 83.9 & 89.0 & 83.7 & 92.8 & 76.3 \\
		\midrule
		\multirow{3}{*}{GPT4} & COFT\_p & 88.2 & 90.0 & 86.5 & 89.7 & 92.6 & 87.0 & 92.6 & 88.4 & 97.3 \\
		 & COFT\_s & 84.1 & 88.1 & 80.5 & 78.7 & 80.5 & 77.0 & 88.3 & 91.2 & 85.5 \\
		 & COFT\_w & 92.4 & 93.3 & 91.5 & 81.4 & 90.7 & 73.9 & 87.8 & 93.8 & 82.5 \\
		\bottomrule
	\end{tabular}
	}
\end{table}

\subsection{Only Input Highlights for LLM Inference}

We incorporate specific symbols to highlight these units within the context to preserve the complete contextual semantics. The absence of complete contextual semantics may face inevitable information loss in real scenarios with more complex attention distributions \cite{hallu_toward_survey}. 

We further conduct experiments on the knowledge hallucination dataset, where we only use the highlighted lexical units as input to LLMs. As shown in Table \ref{tab:only_highlight}, we find that even when only the highlighted lexical units are provided as reference context, the model achieves notable improvements over the baseline methods. This outcome demonstrates the efficacy of our COFT approach in accurately identifying and leveraging support facts within the reference text, thereby enhancing the inference performance. Meanwhile, compared to Table \ref{Tab:hallucination1}, we find that COFT uses only the highlighted lexical units as input is less competitive than the original version of COFT, which also demonstrates the effectiveness of our highlight mechanism.

\begin{table*}[ht]
    \caption{Results of knowledge hallucination benchmark on WK (world knowledge), Sci/Tech (science and technology), and Wri/Rec (writing and recommendation) domains by only making use of the selected key lexical units as input to the LLM. We denote COFT at the paragraph, sentence, and word levels as COFT\_p, COFT\_s, and COFT\_w, respectively.}\label{tab:only_highlight}
    \centering
    \resizebox{\textwidth}{!}{%
        \begin{tabular}{llccccccccccc}
            \toprule
            Backbone & Method & \multicolumn{3}{c}{WK} & \multicolumn{3}{c}{Sci/Tech} & \multicolumn{3}{c}{Wri/Rec} \\
            \cmidrule(lr){3-5} \cmidrule(lr){6-8} \cmidrule(lr){9-11}
             &  & F1 Score & Precision & Recall & F1 Score & Precision & Recall & F1 Score & Precision & Recall \\
            \midrule
            \multirow{3}{*}{Vicuna-33B} 
            & COFT\_p & 69.0 & 69.8 & 68.3 & 61.3 & 57.7 & 65.3 & 65.4 & 60.6 & 71.1 \\
            & COFT\_s & 61.1 & 60.1 & 62.1 & 61.8 & 60.5 & 63.2 & 61.7 & 60.0 & 63.5 \\
            & COFT\_w & 60.1 & 57.1 & 63.5 & 61.9 & 62.8 & 61.1 & 70.1 & 63.1 & 78.8 \\
            \midrule
            \multirow{3}{*}{ChatGPT} 
            & COFT\_p & 74.2 & 80.0 & 69.2 & 77.3 & 78.6 & 76.1 & 70.9 & 76.8 & 65.8 \\
            & COFT\_s & 71.1 & 69.4 & 72.8 & 74.5 & 76.7 & 72.4 & 70.3 & 77.5 & 64.3 \\
            & COFT\_w & 69.7 & 69.9 & 69.5 & 74.3 & 72.3 & 76.5 & 70.9 & 75.5 & 66.9 \\
            \midrule
            \multirow{3}{*}{ChatGPT} 
            & COFT\_p & 77.3 & 75.4 & 79.3 & 86.6 & 80.8 & 93.3 & 85.7 & 80.3 & 91.9 \\
            & COFT\_s & 74.4 & 84.5 & 66.4 & 75.4 & 81.1 & 70.4 & 81.3 & 84.4 & 78.5 \\
            & COFT\_w & 69.9 & 70.1 & 69.8 & 74.3 & 83.5 & 66.9 & 78.6 & 85.1 & 73.0 \\
            \bottomrule
        \end{tabular}
    }
\end{table*}

\subsection{More Results of Analyzing Position Bias of COFT}

We further conduct experiments to explore the impact of position bias in the QA task. Specifically, each reference context comprised 
 relevant documents and 
 irrelevant documents. Drawing upon \cite{lostinmid}, we experiment by varying the positioning of the correct text from the first to the fifth position. As shown in Table \ref{tab:position_bias}, we find that COFT is less influenced by the position bias compared to Vanilla LLMs. This demonstrates COFT's robustness to the positioning of the correct text and implies the great potential to handle lengthy contexts in real-world scenarios.

\begin{table*}[ht]
    \caption{Results of question answering tasks in the Natural Questions benchmark. Performance of the vanilla LLM and COFT is evaluated in terms of EM and F1 score under different positions of the correct document.}
    \label{tab:position_bias}
    \centering
    \resizebox{\textwidth}{!}{%
        \begin{tabular}{l l  c c  c c  c c  c c  c c}
            \toprule
            Backbone & Methods & \multicolumn{2}{c}{1st} & \multicolumn{2}{c}{2nd} & \multicolumn{2}{c}{3rd} & \multicolumn{2}{c}{4th} & \multicolumn{2}{c}{5th} \\
            \cmidrule{3-12}
             & & EM & F1 Score & EM & F1 Score & EM & F1 Score & EM & F1 Score & EM & F1 Score \\
            \midrule
            \multirow{2}{*}{Vicuna-33B}
            & COFT & 21.5 & 32.3 & 20.9 & 32.0 & 19.5 & 30.6 & 20.5 & 31.7 & 21.4 & 32.0 \\
            & Vanilla & 18.3 & 26.3 & 16.3 & 23.6 & 11.7 & 18.6 & 13.3 & 24.5 & 16.2 & 25.8 \\
            \midrule
            \multirow{2}{*}{ChatGPT}
            & COFT & 35.6 & 53.8 & 35.3 & 51.2 & 31.0 & 50.3 & 32.7 & 51.4 & 35.2 & 52.8 \\
            & Vanilla & 28.1 & 40.1 & 26.3 & 37.3 & 20.5 & 32.1 & 23.3 & 33.4 & 26.9 & 38.1 \\
            \midrule
            \multirow{2}{*}{GPT-4}
            & COFT & 58.5 & 66.8 & 57.7 & 66.5 & 56.9 & 64.8 & 57.5 & 65.9 & 57.9 & 65.5 \\
            & Vanilla & 54.5 & 63.5 & 53.7 & 62.7 & 51.6 & 58.4 & 53.6 & 61.1 & 53.9 & 62.8 \\
            \bottomrule
        \end{tabular}
    }
\end{table*}

\subsection{More Results of Randomly Selecting Highlights of COFT}

\begin{table*}[ht]
\caption{Results of knowledge hallucination task. We denote randomly selecting version of COFT as Random COFT and the original version of COFT as Original COFT.}
\label{tab:randomly_selecting}
\centering
\resizebox{\textwidth}{!}{%
\begin{tabular}{llccccccccc}
\toprule
& & \multicolumn{3}{c}{\textbf{WK}} & \multicolumn{3}{c}{\textbf{Sci/Tech}} & \multicolumn{3}{c}{\textbf{Wri/Rec}} \\
\cmidrule(lr){3-5} \cmidrule(lr){6-8} \cmidrule(lr){9-11}
\textbf{Backbone} & \textbf{Methods} & \textbf{F1 Score} & \textbf{Precision} & \textbf{Recall} & \textbf{F1 Score} & \textbf{Precision} & \textbf{Recall} & \textbf{F1 Score} & \textbf{Precision} & \textbf{Recall} \\
\midrule
\multirow{6}{*}{Vicuna-33B} & Random COFT\_p & 59.4 & 67.3 & 53.2 & 50.4 & 51.1 & 49.8 & 55.5 & 50.1 & 62.1 \\
                            & Original COFT\_p & 69.3 & 71.9 & 66.9 & 67.9 & 62.9 & 73.8 & 70.4 & 66.8 & 74.4 \\
                            & Random COFT\_s & 49.9 & 49.7 & 50.1 & 50.9 & 47.9 & 54.3 & 55.1 & 48.9 & 63.1 \\
                            & Original COFT\_s & 62.0 & 63.1 & 60.9 & 68.7 & 67.1 & 70.4 & 66.2 & 64.7 & 67.7 \\
                            & Random COFT\_w & 35.8 & 33.1 & 38.9 & 37.7 & 29.3 & 52.9 & 45.1 & 34.9 & 63.7 \\
                            & Original COFT\_w & 64.4 & 61.7 & 67.4 & 70.9 & 65.7 & 77.2 & 77.3 & 67.9 & 89.8 \\
\midrule
\multirow{6}{*}{ChatGPT}    & Random COFT\_p & 49.5 & 53.8 & 45.9 & 45.8 & 55.4 & 39.0 & 51.0 & 59.2 & 44.8 \\
                            & Original COFT\_p & 78.6 & 83.8 & 74.0 & 83.9 & 81.2 & 86.8 & 77.5 & 85.9 & 70.5 \\
                            & Random COFT\_s & 37.3 & 48.1 & 30.4 & 46.4 & 45.5 & 47.3 & 39.6 & 48.1 & 33.6 \\
                            & Original COFT\_s & 76.8 & 75.7 & 77.9 & 74.6 & 79.1 & 70.5 & 76.8 & 84.4 & 70.5 \\
                            & Random COFT\_w & 52.6 & 53.6 & 51.7 & 57.4 & 49.3 & 68.7 & 47.0 & 55.3 & 40.8 \\
                            & Original COFT\_w & 81.6 & 85.5 & 77.9 & 84.4 & 80.9 & 88.4 & 81.1 & 93.7 & 71.5 \\
\midrule
\multirow{6}{*}{GPT-4}      & Random COFT\_p & 64.1 & 59.1 & 70.0 & 66.0 & 60.1 & 73.3 & 65.0 & 78.9 & 55.3 \\
                            & Original COFT\_p & 83.1 & 79.7 & 86.8 & 89.9 & 84.4 & 96.1 & 91.8 & 85.5 & 99.1 \\
                            & Random COFT\_s & 66.0 & 75.0 & 58.9 & 61.6 & 68.1 & 56.3 & 66.9 & 75.4 & 60.1 \\
                            & Original COFT\_s & 80.0 & 92.3 & 70.6 & 76.6 & 84.9 & 69.8 & 85.5 & 89.2 & 82.1 \\
                            & Random COFT\_w & 62.4 & 80.5 & 51.0 & 71.6 & 77.1 & 66.9 & 68.8 & 78.4 & 61.3 \\
                            & Original COFT\_w & 87.3 & 94.8 & 80.9 & 77.9 & 86.0 & 71.3 & 84.7 & 92.9 & 77.9 \\
\bottomrule
\end{tabular}
}
\end{table*}

We also conduct additional experiments on the knowledge hallucination task. For each query, we randomly highlight lexical units (paragraphs, sentences, or words) that align with the number of highlighted key lexical units in original COFT.

As shown in Table \ref{tab:randomly_selecting}, we find that randomly highlighting lexical units can not improve the results, which demonstrates the effectiveness of COFT to identify key lexical units that are relevant to the query.

\subsection{More Results of Joint-level Highlight Version of COFT}

\begin{table*}[ht]
    \caption{Results of knowledge hallucination benchmark. We denote the joint-level version of COFT as COFT-joint and the original version of COFT as Original COFT. The best results for each LLM backbone are highlighted in \textbf{bold}.}
    \label{Tab:joint_coft}
    \centering
    \resizebox{\textwidth}{!}{
        \begin{tabular}{l l c c c c c c c c c}
            \toprule
            & & \multicolumn{3}{c}{\textbf{WK}} & \multicolumn{3}{c}{\textbf{Sci/Tech}} & \multicolumn{3}{c}{\textbf{Wri/Rec}} \\
            \cmidrule(lr){3-5} \cmidrule(lr){6-8} \cmidrule(lr){9-11}
            \textbf{Backbone} & \textbf{Methods} & \textbf{F1 Score} & \textbf{Precision} & \textbf{Recall} & \textbf{F1 Score} & \textbf{Precision} & \textbf{Recall} & \textbf{F1 Score} & \textbf{Precision} & \textbf{Recall} \\
            \midrule
            \multirow{4}{*}{Vicuna-33B} & Original COFT\_p & {69.3} & {71.9} & 66.9 & 67.9 & 62.9 & {73.8} & 70.4 & 66.8 & {74.4} \\
            & Original COFT\_s & 62.0 & 63.1 & 60.9 & {68.7} & {67.1} & 70.4 & 66.2 & 64.7 & 67.7 \\
            & Original COFT\_w & 64.4 & 61.7 & {67.4} & 70.9 & 65.7 & 77.2 & {77.3} & {67.9} & 89.8 \\
            & COFT-joint & 71.2 & 73.3 & 69.2 & 70.7 & 66.8 & 75.1 & 79.1 & 69.8 & 91.2 \\
            \midrule
            \multirow{4}{*}{ChatGPT} & Original COFT\_p & {78.6} & {83.8} & 74.0 & {83.9} & {81.2} & 86.8 & {77.5} & {85.9} & 70.5 \\
            & Original COFT\_s & 76.8 & 75.7 & {77.9} & 74.6 & 79.1 & 70.5 & 76.8 & 84.4 & 70.5 \\
            & Original COFT\_w & 81.6 & 85.5 & 77.9 & 84.4 & 80.9 & {88.4} & 81.1 & 93.7 & {71.5} \\
            & COFT-joint & 81.1 & 87.1 & 75.9 & 86.2 & 83.5 & 89.0 & 84.8 & 92.5 & 78.3 \\
            \midrule
            \multirow{4}{*}{GPT4} & Original COFT\_p & 83.1 & 79.7 & {86.8} & {89.9} & 84.4 & {96.1} & {91.8} & 85.5 & {99.1} \\
            & Original COFT\_s & 80.0 & {92.3} & 70.6 & 76.6 & 84.9 & 69.8 & 85.5 & 89.2 & 82.1 \\
            & Original COFT\_w & \textbf{87.3} & {94.8} & 80.9 & 77.9 & {86.0} & 71.3 & 84.7 & \textbf{92.9} & 77.9 \\
            & COFT-joint & 89.7 & 95.0 & 85.0 & 90.0 & 88.0 & 92.0 & 92.1 & 90.7 & 93.5 \\
            \bottomrule
        \end{tabular}
    }
\end{table*}

In practical applications, within a document, some paragraphs may be too short and might not require paragraph-level highlightintg, while others may be too long for word-level highlighting. 

 Therefore, we also conduct experiments using a joint-level highlighting version of COFT on the knowledge hallucination task. We highlight key lexical units by word-level granularity. If more than one-third of the words within a sentence were highlighted, we highlight the whole sentence. Similarly, if more than one-third of the sentences within a paragraph were highlighted, we would highlight the whole paragraph.

 As shown in Table \ref{Tab:joint_coft}, COFT at the joint level yields an improvement over using single-level lexical units such as words, sentences, or paragraphs. We will include the joint level version of COFT in Table \ref{Tab:hallucination1} of the main text to have a more comprehensive understanding of COFT. This suggests that exploring more joint-level highlighting strategies could be a promising direction for COFT.

\section{Prompt Templates for Each Task}
We list the prompt templates for different tasks to offer more visually intuitive results in Figures \ref{case_felm}, \ref{case_rc}, and \ref{case_nq} for knowledge hallucination, reading comprehension, and question answering, respectively. More detailed prompt information for the best performance of each task and dataset can be seen within the code.



\clearpage
\section{More Discussions On COFT}
\subsection{Why is it necessary to use both TF-ISF and self-information to measure the importance of entities?}
We propose \textit{context weights} to assess the importance of each entity within a given context. Firstly, we consider the frequency and distribution of entities in the reference context by calculating the \textbf{T}erm \textbf{F}requency-\textbf{I}nverse \textbf{S}entence \textbf{F}requency (TF-ISF), which helps to distinguish entities that are frequently mentioned yet not common words. Such entities play a significant role in understanding the semantics of sentences. Subsequently, we take into account the amount of information an entity contributes when responding to a query within the reference context by computing self-information, which helps to identify potentially significant, highly informative entities. These entities may carry unique or crucial information essential for understanding the entire text.

TF-ISF measures the distinctiveness of words across sentences and self-information assesses their information value in probabilistic terms. By multiplying these two metrics, we obtain a comprehensive indicator, \textit{context weights},  that more effectively captures the importance of words in context. 
Moreover, the \textit{context weights} may potentially be beneficial to consider additional factors or optimize the combination method of TF-ISF score and self-information to get improved results and we leave the exploration as a future work.

\subsection{Why does COFT only search for one-hop neighbor entities in the open-source KG?}

 Retrieving neighbor entities of a target entity in a knowledge graph is a common method for finding entities related to the target. For example, the one-hop neighbor entities of a celebrity often represent attributes like their family, friends, nationality, and workplace, which contain significant and comprehensive information about the given entity. We could also opt to retrieve the two-hop neighbor entities. If the average degree of nodes in the KG is high, this approach can introduce more relevant nodes but may also bring in a large number of irrelevant nodes. 
 
 Therefore, the decision to search for one-hop, two-hop, or more distant neighbor entities should be dynamically adjusted based on the task and the specific characteristics of the KG. For COFT, we observe that retrieving only one-hop neighbors yields satisfactory results. Considering the potential noise introduced by retrieving higher-hop neighbors, we only retrieve one-hop neighbors to enrich the key entity candidates for COFT in all mentioned tasks.




\subsection{Why highlighting key lexical units in three different granularity levels from coarse to fine?}

COFT represents a key entity-driven highlighting approach. It captures potential key entities from the perspective of a knowledge graph (world knowledge) and evaluates the importance of entities based on the TF-ISF and self-information scores. After identifying the final key entities, a straightforward method is to highlight these corresponding entities in the reference context, i.e., the word-level highlighting COFT. However, considering practical applications, such as cases where the reference context contains a small number of entities that appear multiple times, word-level highlighting provides limited information as well. In these scenarios, sentence-level highlighting or paragraph-level highlighting may be more appropriate. 

Moreover, for certain queries, it is crucial to focus on the core paragraphs of the reference text, rather than just sentences or words. Therefore, to enhance the versatility of COFT, we have proposed three different levels of granularity of highlighting selections: paragraph, sentence, and word. Table \ref{Tab:hallucination1} corroborates the effectiveness of multi-granular highlighting as well.

\subsection{What named entity recognition method is employed in COFT, and does it have any tailored designs?}

Given the relative maturity of named entity recognition (NER) in the fields of natural language processing and knowledge graphs, we do not elaborate on it in the main text. Considering the need for rapid deployment and ease of implementation, we have utilized the Spacy\footnote{https://spacy.io/} open-source library for the NER component. Moreover, we employ noun phrase extraction from the NLTK library\footnote{https://www.nltk.org/} to retain some non-named yet significant nouns in queries. We also reference the entity list from Wikidata for entity recognition. 
The ablation study results in Tables \ref{Tab:ab1}, \ref{Tab:ab2}, and \ref{Tab:ab3}, demonstrate the simplicity and effectiveness of the NER component in our method. We also think that other specific optimized NERs are promising to improve COFT.

\end{document}